\documentclass{isprs}
\usepackage{times}
\usepackage{epsfig}
\usepackage{graphics}
\usepackage{multirow}
\usepackage[cmex10]{amsmath}

\usepackage{pgfplots}
\usepackage{pgfplotstable}
\usepackage{url}
\usepackage{setspace}
\usepackage{geometry} 
\usepackage{color}
\usepackage{amssymb}
\usepackage{empheq}
\usepackage{colortbl}
\usepackage{booktabs}
\usepackage[list=true]{subcaption}
\usepackage[labelsep=period]{caption}

\newcommand{\SeeFig}[1]{Figure \ref{#1}}
\newcommand{\SeeSec}[1]{Section \ref{#1}}

\newcommand{\SeeEq}[1]{Equation \ref{#1}}
\newcommand{\SeeTable}[1]{Table \ref{#1}}

\newcommand{\Seefig}[1]{Figure \ref{#1}}

\newcommand{\Seeeq}[1]{Equation \ref{#1}}
\newcommand{\Seetable}[1]{Table \ref{#1}}

\begin{document}

\title{mdBrief - A Fast Online Adaptable, Distorted Binary Descriptor for Real-Time Applications Using Calibrated Wide-Angle Or Fisheye Cameras}

\author{S. Urban and S. Hinz}

\address
{
	Institute of Photogrammetry and Remote Sensing, Karlsruhe Institute of Technology  Karlsruhe \\
	Englerstr. 7, 76131 Karlsruhe, Germany - (steffen.urban, stefan.hinz)@kit.edu \\ http://www.ipf.kit.edu
}

\abstract
{
Fast binary descriptors build the core for many vision based applications with real-time demands like object detection, Visual Odometry or SLAM.
Commonly it is assumed, that the acquired images and thus the patches extracted around keypoints originate from a perspective projection ignoring image distortion or completely different types of projections such as omnidirectional or fisheye.
Usually the deviations from a perfect perspective projection are corrected by undistortion.
Latter, however, introduces severe artifacts if the cameras field-of-view gets larger.
In this paper, we propose a distorted and masked version of the BRIEF descriptor for calibrated cameras.
Instead of correcting the distortion holistically, we distort the binary tests and thus adapt the descriptor to different image regions.

}


\maketitle

\section{Introduction}
\label{ch:features}
The detection and description of salient image features is at the core of most computer vision or photogrammetric processing chains, e.g. object detection, image retrieval, 3D reconstruction, localization, scan registration \cite{urban2015finding},
and numerous methods have been developed over the past decades.
Most features can be grouped into different categories (see \SeeTable{tab:detectors} and \SeeTable{tab:descriptors}) and all have their advantages and disadvantages, depending on the task, the application or the platform.
Commonly, three main steps are involved in feature based algorithms.
\paragraph{Detection}
First the image is searched for keypoints.
Depending on the \textit{detector}, those keypoints can either lie on corners or blob-/region-like structures.
A desirable property of such detectors is that they should be able to repeatably detect the same keypoint over viewpoint changes.
In addition, a robust estimation of the orientation as well as the scale at which the keypoint was detected is necessary for some applications.
\paragraph{Description}
After having identified keypoints, a robust description of the image content surrounding a keypoint is performed.
The description of an image patch is usually mapped into a vector which is called a \textit{descriptor}.
Depending on the method, such descriptors consist either of floating point values or binary strings.
If the keypoint detector provides rotation or scale information, the descriptor can be made invariant to such viewpoint changes.
For many tasks including ego-motion estimation this property is of fundamental necessity as the movement of the camera introduces a continuous change of viewpoint.
\paragraph{Matching}
In the last step, descriptors are matched by comparing their vector representations.
A match is found if the distance between two descriptors is minimal compared to all other descriptors.
Depending on the vector type either the L2-Norm (floating point) or the Hamming distance (binary strings) can be computed. 
Latter has the advantage of speed, as the instruction sets (SSE, AVX) of modern processors allow for parallel bit counting.
In addition, different matching strategies exist. 
One can either match each descriptor from one image to each descriptor from another image, or apply fast approximate nearest neighbor algorithms \cite{muja_flann_2009}.

At this point it is important to emphasize that there does not exist one perfect feature for all tasks.
Some descriptor might outperform all others in terms of recognition rate on one dataset, but might be impractical for real-time applications due to a high computational burden.
Also this paper will not develop a new point feature for omnidirectional images from scratch.
In the first section we will recall the state-of-the-art in image feature research and group it into different categories that are of interest in the context of this work, i.e. developing a real-time capable descriptor for highly distorted images.
The next two sections will extend and adopt recent approaches for the use in online omnidirectional image matching and especially for applications such as SLAM and Visual Odometry.
Finally the last two sections will give results on matching datasets and a conclusion.

\section{State-of-the-Art}
\label{ch:features:sec:stateofart}
This section gives an overview of the state-of-the-art for feature point detectors and descriptors.
As this is a vast topic, we can only cover a small amount of related work.
Instead, this section is supposed to analyze the currently available methods and help to find, adapt and tune a good feature detector - descriptor combination to the given task of egomotion estimation.
For more comprehensive overviews and evaluations of recent research and early developments, the reader is referred to \cite{weinmann2013visual,fan2016local,miksik2012evaluation,gauglitz2011evaluation,Mikolajczyk:pami2005,heinly2012comparative}.

To narrow down the state-of-the-art relevant to the task addressed in this work, lets first analyze the requirements that a detector-descriptor combination for egomotion estimation has to meet.
First of all, the 3-stage process of extraction, description and matching has to be performed in real-time. For a imaging system running at 25 Hz, we have at most 40$ms$ for each image to be processed.
From those 40$ms$ at least 10$ms$ have to be subtracted for pose estimation and other processing steps. 
If there is more than one camera, the remaining 30$ms$ had to be divided by the number of cameras, although there is some room for parallelization.
In addition, the descriptors are supposed to be used in place recognition and re-localization tasks and thus the storage requirements are not negligible as a database of descriptors has to be build.
Apart from the processing speed, the detector-descriptor combination should be invariant (to some extend) to viewpoint and illumination changes, as the motion of a camera system can be unconstrained in terms of rotation and translation.
Latter however allows to relax the requirements on the amount of robustness to significant viewpoint change, as the motion (and thus baseline and rotation) between subsequent frames is rather small and the putative location of the features can be reliably predicted from the last camera poses.
Moreover, the descriptors should be invariant under image distortion or fisheye projections.
As we will see, latter is a requirement rarely addressed.
As todays camera systems are supposed to operate in a changing environment and will visit the same scene more than once, an online adaption of the descriptors that are stored in the database would also be favorable.

\subsection{Keypoint Detectors}
\begin{table*}
	\centering
		\begin{tabular}{| l | c | c | c | c | }
			\hline
			Method & Detector & Blob/Corner & Scale/Orientation/Distortion & Time $\mu s$/Point\\ \hline
			AGAST & AGAST & corner & -/-/- & 0.48 \\ \hline
			FAST & FAST & corner & -/-/- & 0.84 \\ \hline
			ORB & FAST on pyramid & corner & \checkmark/\checkmark/- & 19 \\ \hline
			BRISK & AGAST on pyramid & corner & \checkmark/\checkmark/- & 24 \\ \hline
			SPHORB & SFAST & corner & \checkmark/\checkmark/\checkmark &  $>$ ORB \\ \hline
			SURF & SURF & blob & \checkmark/\checkmark/- & 47 \\ \hline
			A-KAZE & A-KAZE & blob & \checkmark/\checkmark/- & 152 \\ \hline
			SIFT & DoG & blob & \checkmark/\checkmark/- & 170 \\ \hline
			sRD-SIFT & distorted DoG & blob & \checkmark/\checkmark/\checkmark & $>$ SIFT \\ \hline
		\end{tabular}
	\caption{Overview of state-of-the-art keypoints detectors. Except for sRD-SIFT and SFAST, all algorithms are part of OpenCV 3.1.}
	\label{tab:detectors}
\end{table*}
In general, keypoint detectors can be grouped into two categories, i.e. corner and blob-like detectors.
\paragraph{Corners} Two of the first corner detectors, still in use in many applications, are the Harris corner \cite{harris1988combined} and Shi-Tomasi's "Good Features to Track" \cite{shi1994good} detector. 
In both methods, a "corner score" is derived by analysis of the eigenvalues of the second-order moment matrix which involves computing the image derivatives.
To avoid such costly computations FAST \cite{rosten2006machine,rosten2010faster} uses simple gray value comparisons on discretized circle segments around a candidate pixel to decide whether a pixel is classified as a corner.
To speed up the process, they involve machine learning techniques (a ternary Decision Tree) to reduce the number of necessary pixel comparisons to a minimum.
The drawback is that this Decision Tree should be relearned for new environments or scene structures.
Building upon the idea of FAST, the AGAST \cite{Mair2010c} detector improves the segment testing by finding an optimal binary Decision Tree that adapts to new environments.
In addition, the ternary test configuration of FAST (similar, brighter, darker) is extended to be more generic.
So far, none of the detectors is invariant to scale changes or estimates a salient orientation.
The ORB\cite{rublee2011orb} detector builds an image pyramid and extracts FAST features on every level.
To avoid large responses along edges, the authors use the Harris corner score on every keypoint.
In addition, the magnitude of the corner measure is used to order the keypoints.
However, no non-maxima suppression is performed between different scale levels, leading to potential duplicate detections.
To estimate a salient orientation, an intensity centroid is computed in the patch. 
Then, the orientation is defined by the vector between the patch center and this centroid.
The BRISK \cite{leutenegger2011brisk} detector tries to improve upon ORB by extracting AGAST corners on different levels of the image pyramid and scale interpolation between octaves. 
In addition, a non-maxima suppression across scales is performed, using the FAST score as a measure of saliency.
The orientation is estimated from the gradients of long-distance pairs of a distinct sampling pattern around the keypoint that is later also used to calculate the descriptor.

\paragraph{Blobs} Blobs are usually detected as the local extrema of different filter responses.
The SIFT \cite{Lowe_2004} detector generates a scale space by subsequent filtering of the image with separable Gaussian filters with increasing variance.
Then, extrema are detected in Differences of these Gaussian filtered images.
To speed up the computation, the SURF detector \cite{bay2006surf} approximates the Gaussian derivatives by box filters, thus circumventing the creation of the whole Gaussian scale space by simple upscaling of the filter size. 
In addition, both detectors estimate a patch orientation.
SIFT builds a histogram of gradient orientations to select the one that is dominant in the patch.
SURF uses the sums of Gaussian weighted Haar Wavelet responses in two directions to assign an orientation to the keypoint.
The authors of the KAZE \cite{alcantarilla2012kaze} detector as well as its accelerated pendant A-KAZE \cite{alcantarilla2011fast} argue that a linear Gaussian scale space smooths noise and object details to the same degree.
Hence, they build a non-linear scale space using non-linear diffusion filtering to obtain keypoints that exhibit a much higher repeatability.
The keypoint orientation is found using a method similar to the SURF detector.

So far, none of the presented detectors considers image distortion or different projections.
Both introduce deformations to corners and blobs that will decrease the repeatability of the detector. 
In theory, the image could be warped to match a perfect planar perspective projection.
This however introduces severe artifacts and is limited to images with less than a hemispherical view of the scene and should be avoided \cite{daniilidis2002image}.
Thus, all image processing operations have to be carried out directly in the distorted image.
To detect repeatable keypoints in omnidirectional images, invariant to scale and rotation, the construction of the scale space \cite{puig2011scale,hansen2008spherical,bulow2004spherical} as well as the image gradient computation have to be adjusted \cite{furnari2015generalized}. 
Prominent enhancements of the SIFT blob detector are pSIFT \cite{hansen2008spherical} and sRD-SIFT \cite{lourencco2012srd}.
They excel their "standard" pendants regarding matching performance but the adapted construction of scale-spaces on the sphere increases their extraction time significantly and makes them even less suitable for real-time applications.

The same issues occur with corner detectors.
FAST, for example, takes N pixels of a geometric shape (a Bresenham circle) around each candidate keypoint and compares their intensities.
This works well, assuming that the pixel neighborhood remains the same over the image domain.
Certainly, if the image is affected by strong distortion, the pixel neighborhood varies depending on position from the image center.
Thus, the circle had to be adjusted (distorted) depending on the position of the point in the image and pixel intensities would have to be interpolated.
This, however, destroys the high efficiency of the detector.
For SPHORB (Spherical ORB) \cite{zhao2015sphorb}, the authors introduce the geodesic grid, borrowing ideas from climate modeling.
Basically, the sphere is subdivided into Voronoi cells of similar size.
Now, each pixel can be assigned to Voronoi cells and the neighborhood can be directly observed on the sphere and is independent of the position in the image.
They train a FAST detector on the spherical neighborhood representation and dub the detector SFAST (Spherical FAST).

Summarizing, most detectors satisfy the requirements on providing a rotation and scale estimate (\Seetable{tab:detectors}).
Blob-like detectors, though, remain computationally to expensive for real-time applications.
BRISK and ORB, however, are a lot faster than their blob-like competitors by adapting highly efficient corner detectors in image pyramid schemes.
Concerning robustness to image distortions and non-perspective projections, the range is limited.
The proposed blob detectors are even slower than their predecessors.
SFAST could be an alternative, but at the time of writing was not available.

In the remainder of this work, the ORB detector scheme is used to detect keypoints.
It is still the fastest method providing a stable rotation estimate.
By replacing the FAST detector with AGAST, we are able to gain some additional $ms$ and do not have to retrain the detector to different environments. 
Yet, the aforementioned drawbacks of using the FAST (or AGAST) detector on fisheye images remains and could be subject to future work.

\subsection{Keypoint Descriptors}
\begin{table*}
	\centering
		\begin{tabular}{| l | c | c | c | c | c | c |}
			\hline
			Descriptor & type & scale/rotation/distortion & \#byte & speed $\mu s$/point & off-line learned & online adaption \\ \hline
			BRIEF & (IV) binary & \checkmark/\checkmark/- & 16-64 & 6 & - & - \\ \hline
			ORB & (IV) binary & \checkmark/\checkmark/- & 32 & 14 & \checkmark & - \\ \hline
			BRISK & (IV) binary & \checkmark/\checkmark/- & 64 & 20 & - & - \\ \hline	

			BOLD & (IV) binary & \checkmark/\checkmark/- & 64 & 88 & \checkmark & \checkmark \\  \hline
			BinBoost & (IV) binary & \checkmark/\checkmark/- & 8 & 420 & \checkmark & - \\ \hline
			TailoredBRIEF & (IV) binary & \checkmark/\checkmark/- & 16-64 & - & - & \checkmark \\ \hline
			SPHORB & (IV) binary & \checkmark/\checkmark/\checkmark & 32 & - & \checkmark & - \\ \hline
			SURF & (II) real-valued & \checkmark/\checkmark/- & 256 & 76 & - & - \\ \hline
			SIFT & (II) real-valued & \checkmark/\checkmark/- & 512 & 239 & - & - \\ \hline
			sRD-SIFT & (II) real-valued & \checkmark/\checkmark/\checkmark & 512 & - & - & - \\ \hline
		\end{tabular}
	\caption{Overview of state-of-the-art descriptors.}
	\label{tab:descriptors}
\end{table*}
To facilitate the choice of a descriptor for any given task, it is usually advantageous to classify given methods into different categories that are of interest.
In \cite{heinly2012comparative} the authors propose a taxonomy that puts descriptors into four categories relating the time to compute a descriptor and its storage requirements. 
Both are also important in the context of large-scale egomotion estimation and we adopt their taxonomy.
In the following, we assume that a detector was used to extract keypoints with corresponding orientation and scale information.

\begin{figure*}
	\centering
	\subcaptionbox{\label{fig:briefPatternRandom}}
	{\includegraphics[width=0.48\textwidth]{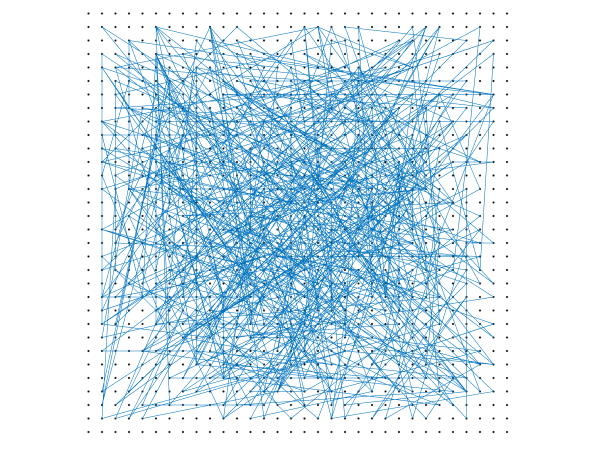}}
	\subcaptionbox{\label{fig:briefPatternLearned}}
	{\includegraphics[width=0.48\textwidth]{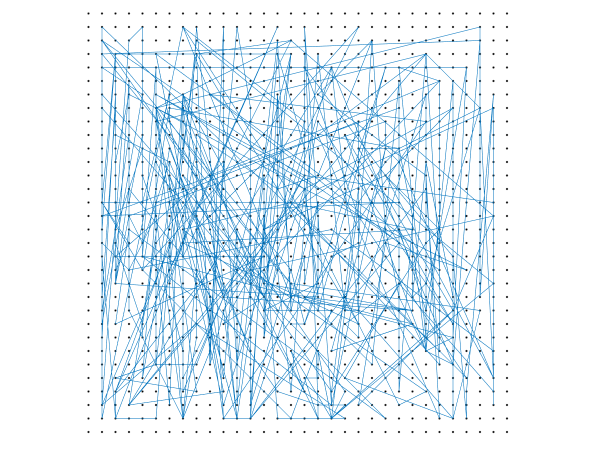}}
	\caption{Depicted are 256 tests of a test set $\mathbf{Q}$ (a) The tests are randomly sampled from an isotropic Gaussian distribution, mirroring the BRIEF descriptor. (b) The tests are learned in a scheme similar to ORB. The tests have the property of low correlation and high variance.} 
	\label{fig:briefPattern}	
\end{figure*}

\paragraph{Type I - Image patches} 
The most basic type of descriptor is an image patch surrounding the detected keypoint. The matching of such image patches can then be performed by computing, for example, the sum of squared differences (SSD) or the normalized cross correlation (NCC).
Scale, rotation and distortion invariance can be achieved, by rotating, scaling and undistoring the patch content which, however, increases the computational burden per patch.
In addition, the storage requirements grow quadratically with the patch size, making this type of descriptor impracticable for large scale operations.
\paragraph{Type II - Real-valued}
The second category contains real-valued descriptors such as SIFT and SURF. 
By applying a succession of image processing operations such as image derivatives, linear transformations, spatial pooling, histograms, etc. the patch content is mapped to a vector of fixed length reducing the memory footprint.
In this category, rotation invariance is achieved by rotating gradient orientations w.r.t a dominant orientation and scale can be compensated by computing the descriptor on the image corresponding to the keypoint scale in the image pyramid.
Compared to the type I descriptors, however, the computational effort grows as each patch undergoes the series of image processing steps.
\paragraph{Type III - Binarized} The algorithms in the third category try to reduce the memory footprint even further by means of quantizing the real valued descriptors into binary strings (binarization) \cite{calonder2009compact,winder2009picking} or by applying dimensionality reduction techniques (PCA-SIFT \cite{ke2004pca}) without significant loss in matching accuracy.
But still, the elaborate pre-processing of the patch content has to be performed and thus the description time for each patch remains high.
\paragraph{Type IV - Binary} This category subsumes so called binary descriptors.
Here, each descriptor dimension is the result of a simple binary comparison on pixel intensities in the patch surrounding the descriptor.
Thus, the time needed to compute a descriptor drops significantly.
This idea was initially proposed by \cite{calonder2010brief} and is called BRIEF.
They define a test $\tau$ on a smoothed image patch $\mathbf{P}$ as follows:
\begin{equation}
\tau(\mathbf{P};\mathbf{u}_1, \mathbf{u}_2)=\left\{\begin{array}{ll} 1, & if  I(\mathbf{p},\mathbf{u}_1) < I(\mathbf{p},\mathbf{u}_2) \\
0, & otherwise \end{array}\right .
\label{eq:testBrief}
\end{equation}
where $I(\mathbf{P},\mathbf{u})$ is the intensity at pixel location $\mathbf{u} = (u,v)^T$ in the patch $\mathbf{P}$.
The set $\mathbf{Q}$ of $d=1,..,D$ test pairs 
\begin{equation}
\mathbf{Q} = 
\begin{pmatrix}	
\mathbf{u}_{11},...,\mathbf{u}_{1d} \\
\mathbf{u}_{21},...,\mathbf{u}_{2d} \\
\end{pmatrix}
= (\tau_1,..,\tau_d)
\label{eq:testSet}
\end{equation}
is fixed and its choice affects the recognition performance.
Calonder et al. experimented with different methods to choose the test locations
and sampling the test set from an isotropic Gaussian distribution worked best.
\SeeFig{fig:briefPatternRandom} depicts such a set of randomly sampled tests.
The size of the descriptor (equivalent to the number of tests) can be arbitrarily chosen.
Naturally, a larger number of tests increases the matching performance but the improvement saturates already around 512 tests \cite{calonder2010brief}.
The descriptor can be made invariant to rotation and scale by simply transforming (rotating, scaling) the tests around the keypoint (patch center).
Unfortunately, the rotation of the test pattern decreases the variance, which makes the feature less discriminative \cite{rublee2011orb}.
In addition, the authors of ORB show that tests can be correlated, i.e. some tests do not contribute equally to the results.
They propose an unsupervised learning scheme to find the tests (of all possible tests in a patch) that have the highest variance, whilst being uncorrelated.
Such a set of learned tests is depicted in \Seefig{fig:briefPatternLearned}.
We will come back to this training scheme in the next section.
Instead of sampling tests randomly or learning uncorrelated test, the BRISK \cite{leutenegger2011brisk} descriptor is build from a set of equally spaced tests that are located on circles concentric with the keypoint.
As mentioned in the detector section, the patch orientation is estimated using the gradients from the long-distance pairs, while the short-distance pairs are used to compute the descriptor.
\paragraph{Learned descriptors}
Thus far, most descriptors are "hand-crafted", i.e. the image processing operations within the patch, as well as their order, were carefully selected and tuned.
Large ground-truth datasets of matching and non-matching patches \cite{winder2009picking} allowed to take a different approach and involve machine learning to map the patch content to a discriminative descriptor.
In \cite{winder2009picking} optimal descriptor parameters for the DAISY \cite{tola2008fast} descriptor are learned.
In \cite{lategahn2013learn} the image processing operations (called building blocks), as well as their order is learned.
To create the BinBoost \cite{trzcinski2013boosting} descriptor, the authors learn a hash function that directly maps patch gradient responses to a binary string.
Although the descriptors of those methods are the results of learning parts of the processing chain, the authors still selected which building blocks to consider.
The recent success of CNNs and their ability to learn which features and representations are relevant to minimize a given cost function, motivated the authors of  to learn not only the "patch to descriptor" mapping \cite{balntas2016pn,wlrn}, as a whole, but also distance metric between two descriptors \cite{simo2015discriminative,han2015matchnet}.


Still, the mapping from patch to descriptor is learned off-line from large datasets of patches, that originate from scenes representing distinct and fixed characteristic.
An intuitive improvement is to adapt the descriptor to new scenes.
Unfortunately, the off-line training involved in most methods is cumbersome and requires (up to) days for training. 
Recent methods \cite{richardsontailoredbrief,balntas2015bold} propose to adapt the BRIEF descriptor presented above for every patch or point, respectively.
The basic idea is to learn a boolean mask of the same length as the fixed binary test set $\mathbf{Q}$ (\Seeeq{eq:testSet}), that suppresses tests that have a varying outcome for different views and are thus unstable.
This mask can be found by transforming the fixed tests $\mathbf{Q}$ on-line and reapply it to the patch.
The resulting descriptors should be equal, if all tests have the same outcome. 
If not, the test will be suppressed during matching. 
For TailoredBRIEF \cite{richardsontailoredbrief} the authors propose to sample many affine views of the tests for reference points online.
Then, while matching descriptors from the current camera image to descriptors from previously extracted landmark points, the learned mask is applied to discard unstable tests.
They mention that masks for both the reference and the current descriptors could be learned, but argue, that for real-time tracking systems this is not necessary.
For BOLD \cite{balntas2015bold}, the authors point out the similarities of learning masks to LDA based approaches \cite{strecha2012ldahash}.
In principal, LDA tries to find a linear combination of features, that best separates two classes (matching, non-matching points), by maximizing inter-class (different patch) and minimizing intra-class (same patch) variance.
Learning uncorrelated tests off-line, like done for the ORB descriptor, increases the inter-class distance by maximizing the variance across inter-class patches.
Then, by sampling and finding stable tests online, the intra-class variance is minimized.
For BOLD, masks for all patches are learned and applied during matching.
We will embrace both ideas (off-line and on-line learning), discuss them in more detail and adapt them to the task of creating a distorted version of the BRIEF descriptor for calibrated wide-angle and fisheye cameras.

Apart from being rotation and scale invariant, the descriptor should be binary, as the storage requirements as well as the matching speed significantly outperform their real-valued counterparts.
Methods that binarize real-valued descriptors (type III) or learn a mapping from patch to descriptor are still to slow for mobile real-time applications.
This leaves us with binary descriptors that are created from binary intensity comparisons.
In addition, the online adaption of such binary descriptors is possible \cite{richardsontailoredbrief,balntas2015bold}.
Ultimately, the question remains how to adapt the binary descriptors to image distortion or non-perspective projections.
Most state-of-the-art descriptors assume that the patch content surrounding a keypoint originated from a perspective projection and can be locally approximated by an affine transformation \cite{morel2009asift}.
Most commonly the lens distortion is ignored, as it is assumed that images are undistorted before feature extraction is performed.
As discussed above, the undistortion of wide-angle, fisheye or omnidirectional images often leads to re-sampling artifacts due to interpolation deteriorating the matching performance. 

In the following section, we will take a closer look at combining the advantages of learning BRIEF-like tests off-line (ORB) and on-line (BOLD) and applying image distortion to the tests instead of the patches, which is a lot faster to compute.

\section{dBRIEF - Distorted BRIEF}
\label{ch:features:sec:distortedbrief}
At the beginning of this section a modified (distorted) version of the BRIEF descriptor is presented.
We then show in simulations how the modifications transfer to Hamming distance distributions between matching and non-matching descriptors and the recognition rate respectively.
Subsequently, an optimal, low correlation test set is learned, following the unsupervised learning scheme presented in \cite{rublee2011orb}.
We contribute an additional study on how the learned test set changes if different detectors are used to acquire the initial set of keypoints for learning.
Further, we investigate if training on fisheye images improves performance.
Following the off-line learning, we present the online mask learning for our distorted descriptor and evaluate the developed methodology against state-of-the-art methods w.r.t speed and matching performance on real data.

To show both the effect of radial distortion as well as non-perspective mappings on the BRIEF descriptor, we first introduce the simple radial distortion model \cite{lourencco2012srd} with one parameter $\xi$ controlling the amount of radial distortion:
\begin{equation}
\mathbf{m}_d = r(\mathbf{m})= \frac{2\mathbf{m}}{1+\sqrt{1-4\xi \|\mathbf{m}\|^2}}
\label{eq:radialDistModel}
\end{equation}
where $\mathbf{m} = [u,v]^T $ is an image point in sensor coordinates and $\mathbf{m}_d$ its distorted counterpart.
Note, that we chose this model for the sake of simplicity and speed, as it is analytically invertible.
This model could be exchanged with every other distortion model.
\begin{figure*}
\hspace{0.22\textwidth}%
\subcaptionbox{\label{fig:undistortionScheme}}%
{\includegraphics[clip=true, trim = 50mm 1mm 50mm 50mm,width=0.5\textwidth]{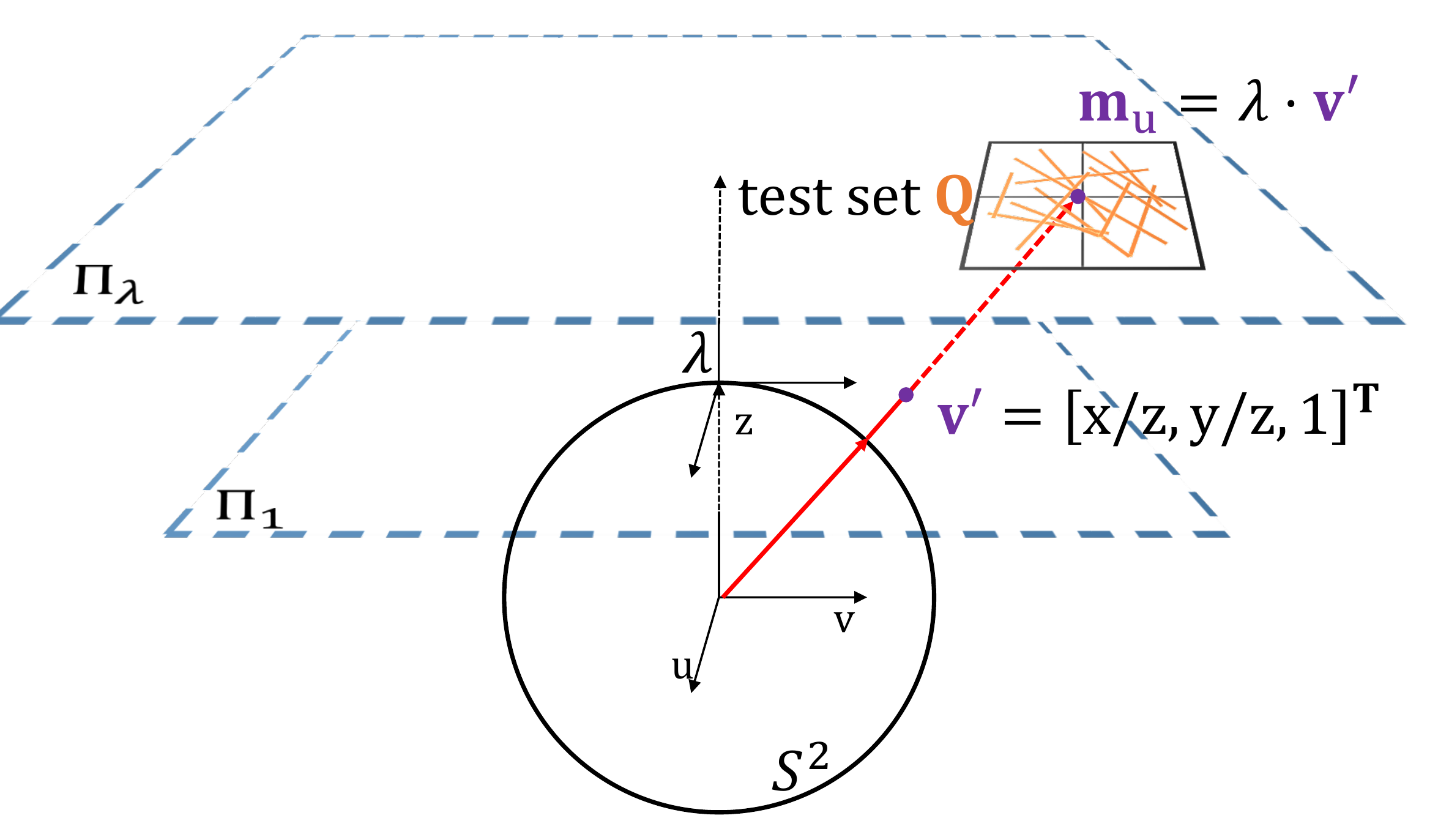}}\\
\subcaptionbox{\label{fig:distPerspectiveScheme}}%
{\includegraphics[clip=true, trim = 50mm 1mm 50mm 50mm,width=0.5\textwidth]{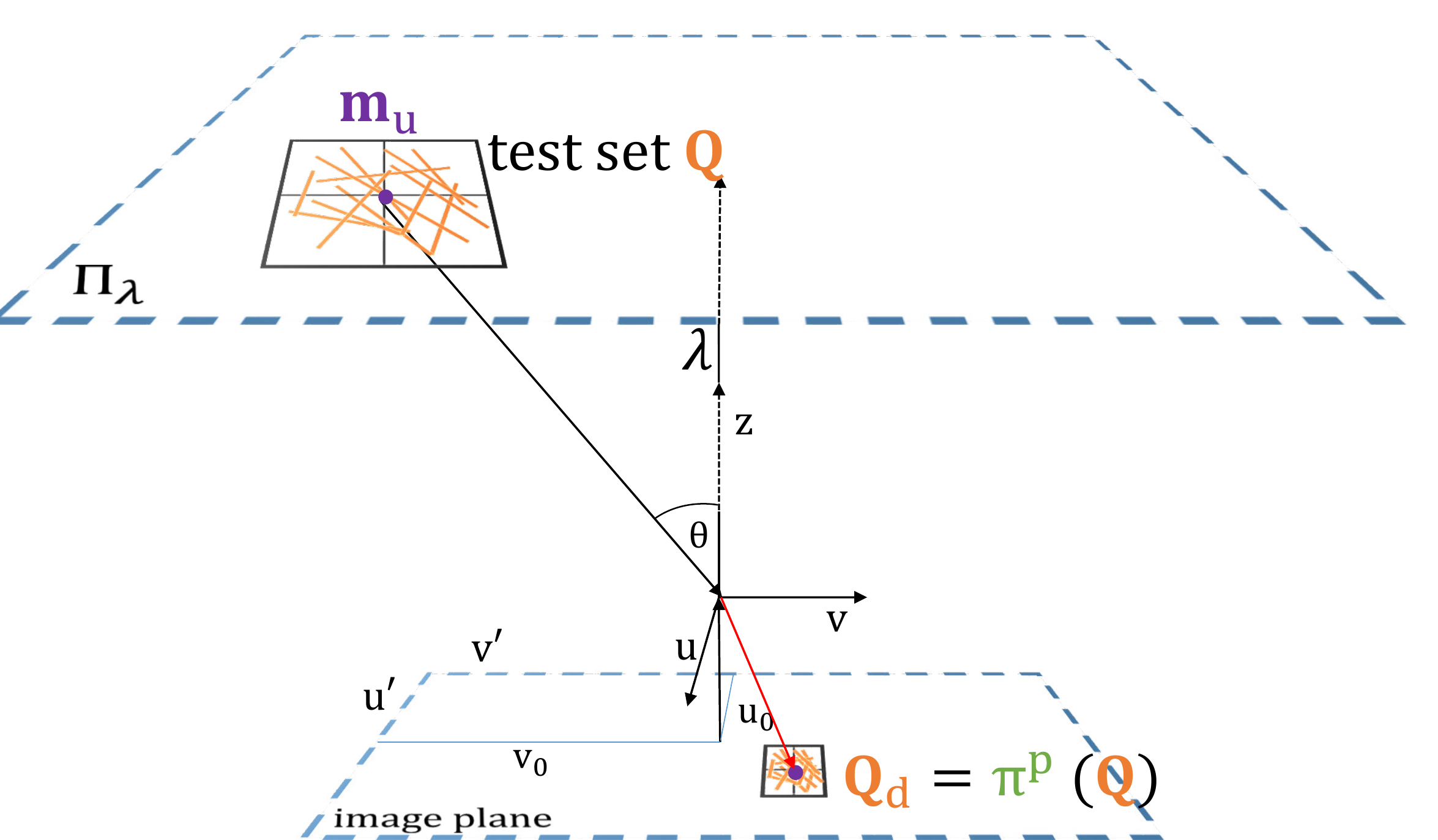}}%
\subcaptionbox{\label{fig:distFishScheme}}%
{\includegraphics[clip=true, trim = 50mm 1mm 50mm 50mm,width=0.5\textwidth]{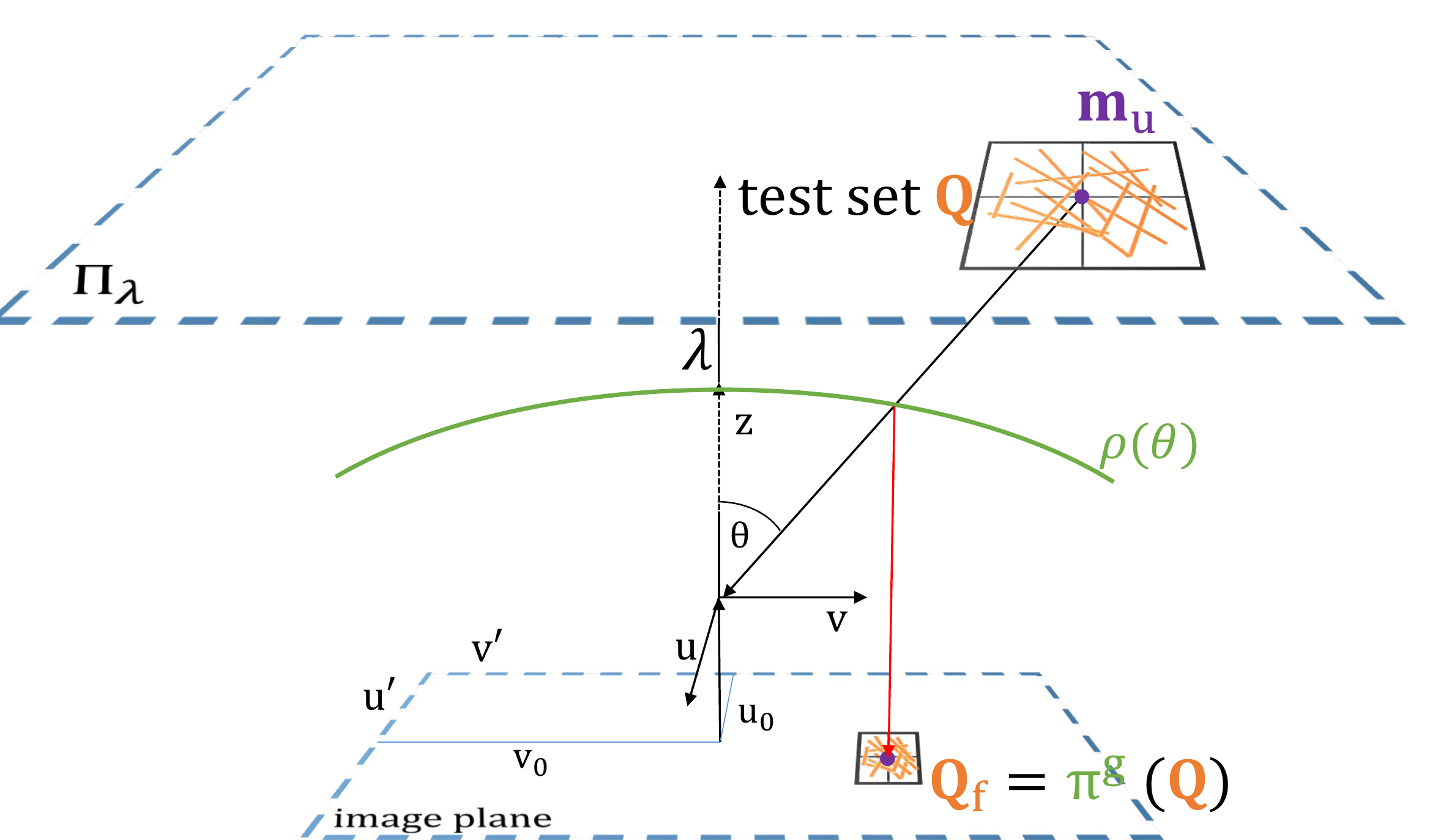}}
\caption{(a) Undistortion of the current image point. After projecting the point to the unit sphere $S^2$ and subsequent projection onto the normalized image plane $\Pi_1$ the point $\mathbf{v}^{'}$ is scaled to a plane $\Pi_{\lambda}$ that lies in distance $\lambda$ from $\Pi_1$. Then the test set $\mathbf{Q}$ is added to the point coordinates.} 
\label{fig:briefDistortionScheme}	
\end{figure*}

From \Seeeq{eq:radialDistModel} it is obvious that the distorted coordinates $\mathbf{m}_d$ depend on the undistorted point $\mathbf{m}$.
In order to calculate a distorted version of our test set $\mathbf{Q}$, we thus have to get the coordinates of the undistorted points first.
Let $\mathbf{m}'$ be a keypoint in image coordinates detected in the distorted image.
Assume an arbitrary interior orientation parametrization $\pi$ including \SeeEq{eq:radialDistModel} then the projection of $\mathbf{m}'$ to its corresponding (undistorted) bearing vector is:
\begin{equation}
\mathbf{v} = [v_x,v_y,v_z] = \pi^{-1}(\mathbf{m}')
\label{eq:undist}
\end{equation}
and its counterpart on the normalized image plane $\Pi_1$:
\begin{equation}
\mathbf{v}' = [v_x/v_z,v_y/v_z,1]
\label{eq:undistNormalization}
\end{equation}
This transformation is depicted in \Seefig{fig:undistortionScheme}.
Now, $\mathbf{v}'$ can be scaled to a plane $\Pi_{\lambda}$ with distance $\lambda$ from the normalized image plane, yielding 
\begin{equation}
\mathbf{m}_u = [m_x,m_y,m_z]^T = \lambda\mathbf{v}'
\label{eq:undistEbeneOben}
\end{equation}
For a perspective camera $\lambda$ would be set to the focal length.
Using a camera model for fisheye and omnidirectional cameras \cite{urban2015improved,scaramuzza2006b}, we can set $\lambda = a_0$, i.e. the first coefficient of the forward projection polynomial.
Subsequently, the fixed test set $\mathbf{Q}$ can be anchored on $\mathbf{m}_u$.
\begin{equation}
\mathbf{Q}_u = \mathbf{Q}+\mathbf{m}_u
\label{eq:undistTesSet}
\end{equation}
The final distorted version of the test set $\mathbf{Q}_d$ is then given by back-projection of $\mathbf{Q}_u$ onto the image plane.
Three realizations, of this projection are given by the following equations:
\small
\begin{empheq}[]{align}
	\text{persp.: }
	\mathbf{Q}
	&=
	\pi^p(\mathbf{p}_{i})
	=
	\frac{\mathbf{Q}_u}{m_z} + \mathbf{o} 	
	\label{eq:classCentral}
	\\
	\text{persp. + dist.: }
	\mathbf{Q}_d
	&=
	r(\frac{\mathbf{Q}_u}{m_z}) + \mathbf{o}
	\label{eq:classCentralRad}
	\\
	\text{fisheye: }
	\mathbf{Q}_f
	&=
	\pi^g(\mathbf{Q}_{u})
	=
	\mathbf{A}
	\begin{bmatrix}
		\rho(\theta)m_x/r \\ \rho(\theta)m_y/r
	\end{bmatrix}
	+
	\begin{bmatrix}
		o_u\\o_v
	\end{bmatrix}
	\label{eq:Omni}		
\end{empheq}
with $r=\sqrt{m^2_x+m^2_y}$ and $\theta=\arctan({m_z/r})$.
\normalsize
The first version (\Seeeq{eq:classCentral}) equals the standard version of the descriptor generation.
In the second version (\Seeeq{eq:classCentralRad}), radial distortion is applied to the tests in the sensor plane using \Seeeq{eq:radialDistModel}. 
This projection is depicted in \Seefig{fig:distPerspectiveScheme}.
The third version is the non-perspective projection of the test set.
We use the generic camera model introduced by \cite{scaramuzza2006b}, but other projection models could be used as well \cite{schneider2009validation}.

\begin{figure*}
	\centering
	\begin{minipage}[t]{\textwidth}
		\centering 		
		{\fcolorbox{red}{white}%
		{\includegraphics[width=0.15\textwidth]{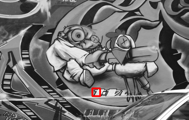}%
		\includegraphics[width=0.15\textwidth]{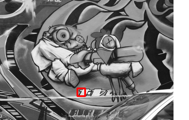}%
		\includegraphics[width=0.15\textwidth]{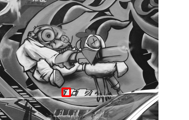}%
		\includegraphics[width=0.15\textwidth]{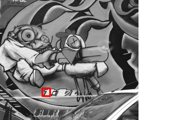}%
		\includegraphics[width=0.15\textwidth]{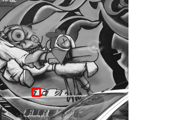}%
		\includegraphics[width=0.15\textwidth]{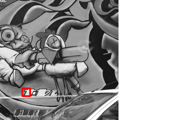}}\hfill%
		\fcolorbox{blue}{white}%
		{\includegraphics[width=0.15\textwidth]{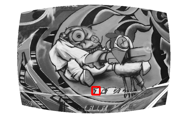}%
		\includegraphics[width=0.15\textwidth]{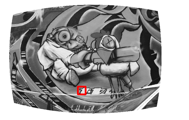}%
		\includegraphics[width=0.15\textwidth]{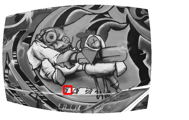}%
		\includegraphics[width=0.15\textwidth]{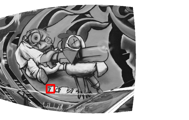}%
		\includegraphics[width=0.15\textwidth]{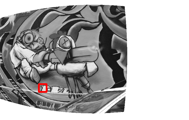}%
		\includegraphics[width=0.15\textwidth]{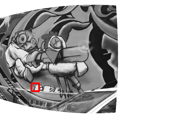}}\hfill%
		\fcolorbox{green}{white}%
		{\includegraphics[width=0.15\textwidth]{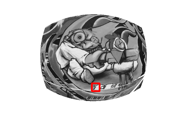}%
		\includegraphics[width=0.15\textwidth]{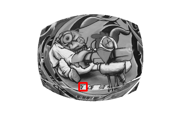}%
		\includegraphics[width=0.15\textwidth]{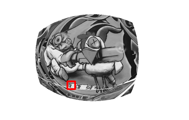}%
		\includegraphics[width=0.15\textwidth]{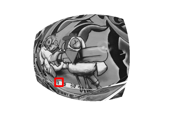}%
		\includegraphics[width=0.15\textwidth]{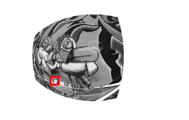}%
		\includegraphics[width=0.15\textwidth]{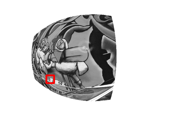}}}
		\caption{Tracking simulation. Camera is moving in X-direction.Red box: perspective projection. Blue box: added radial distortion ($r_d=-2^{-6}$). Green box: fisheye projection.} 
		\label{fig:simulationXtracking}	
	\end{minipage}\hfill
	\begin{minipage}[t]{\textwidth}
		{\includegraphics[width=\textwidth]{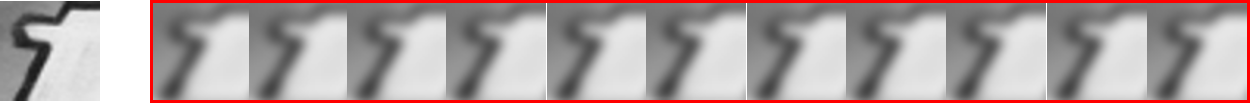}}\hfill 
		{\includegraphics[width=\textwidth]{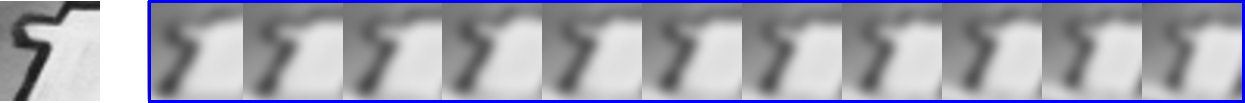}}\hfill
		{\includegraphics[width=\textwidth]{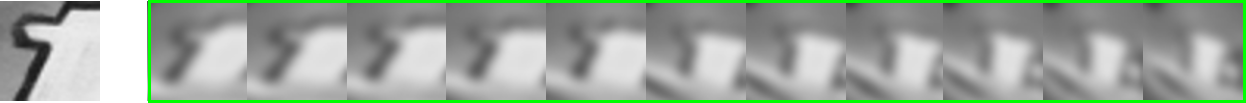}}
		\caption{Patch (red square in \Seefig{fig:simulationXtracking}) for each of the three cases. Again: red for perspective projection, blue for radial distortion, green for fisheye projection.}
		\label{fig:patchesCorrespondingTracking}
	\end{minipage}\hfill
	\begin{minipage}[t]{\textwidth}
		\centering  
		\fcolorbox{red}{white}{
			{\includegraphics[width=0.15\textwidth]{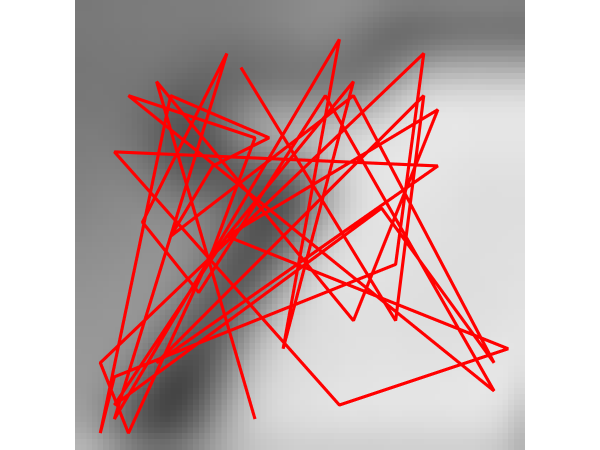}
				\includegraphics[width=0.15\textwidth]{grafImg_off_patch_perspective_tests1.png}
				\includegraphics[width=0.15\textwidth]{grafImg_off_patch_perspective_tests1.png}
				\includegraphics[width=0.15\textwidth]{grafImg_off_patch_perspective_tests1.png}
				\includegraphics[width=0.15\textwidth]{grafImg_off_patch_perspective_tests1.png}
				\includegraphics[width=0.15\textwidth]{grafImg_off_patch_perspective_tests1.png}} 
		}	
		\fcolorbox{blue}{white}{
			{\includegraphics[width=0.15\textwidth]{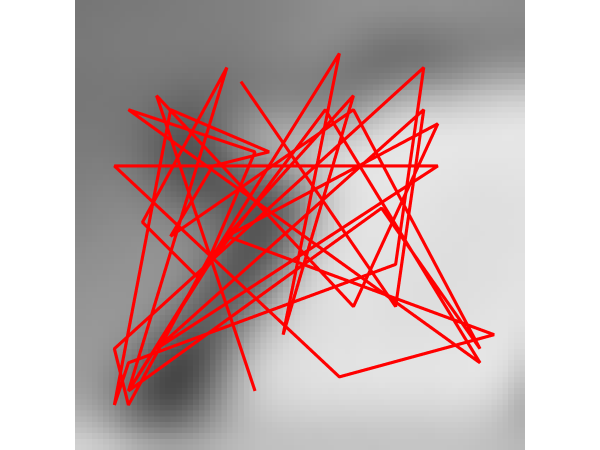}
				\includegraphics[width=0.15\textwidth]{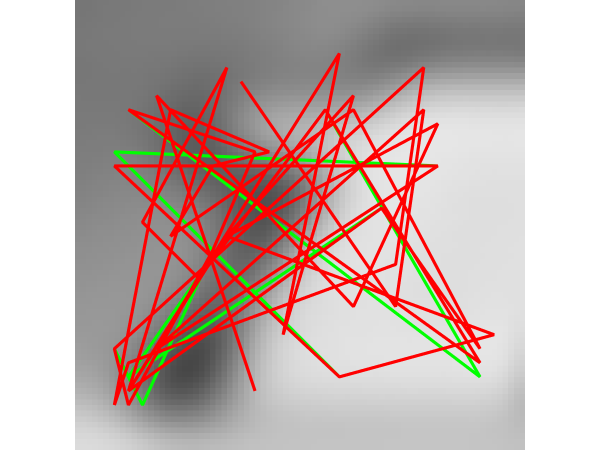}
				\includegraphics[width=0.15\textwidth]{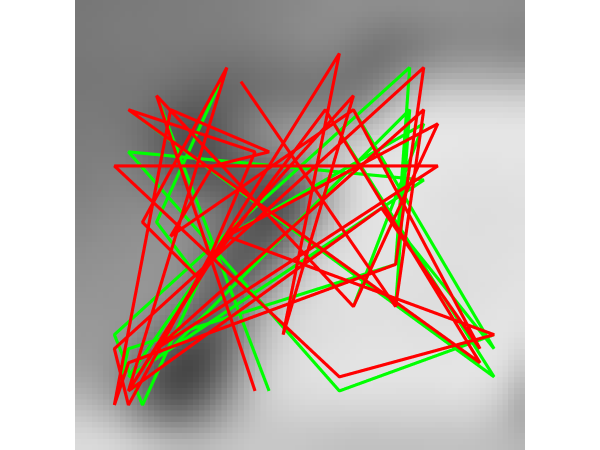}
				\includegraphics[width=0.15\textwidth]{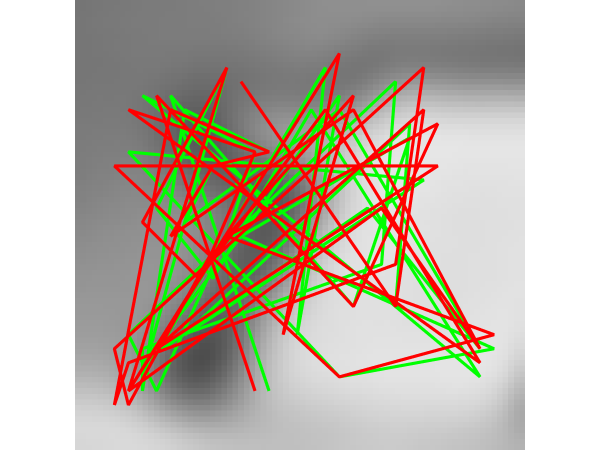}
				\includegraphics[width=0.15\textwidth]{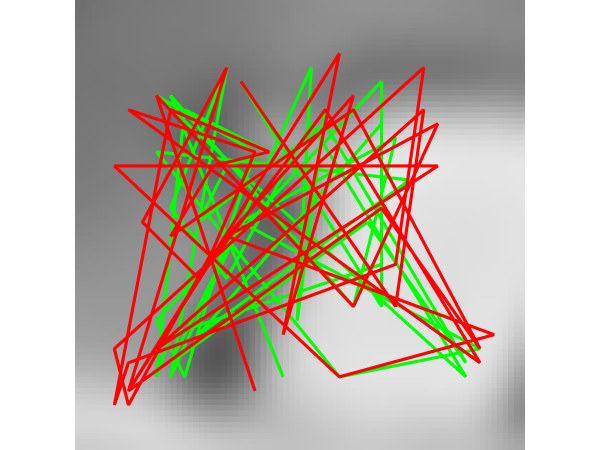}
				\includegraphics[width=0.15\textwidth]{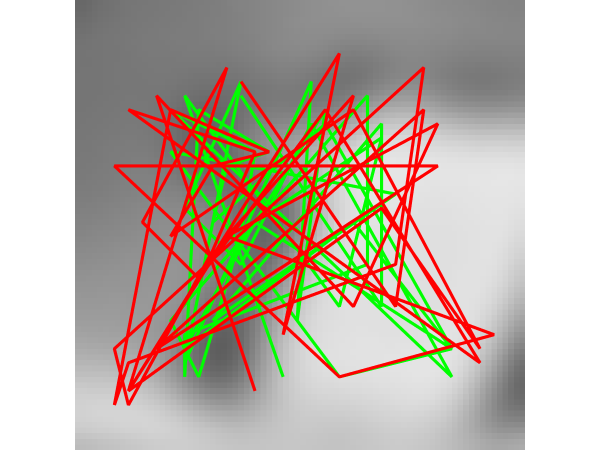}} 
		}
		\fcolorbox{green}{white}{
			{\includegraphics[width=0.15\textwidth]{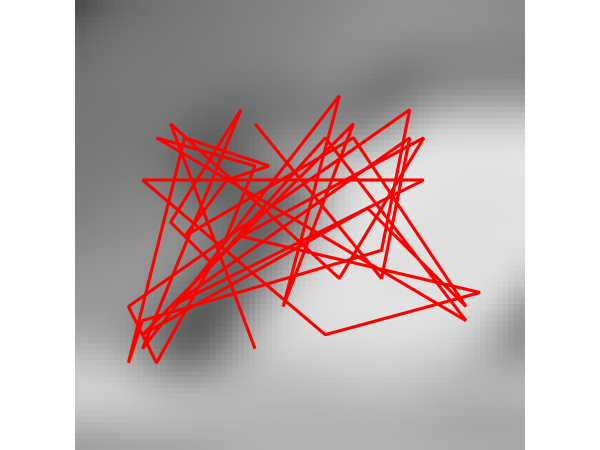}
				\includegraphics[width=0.15\textwidth]{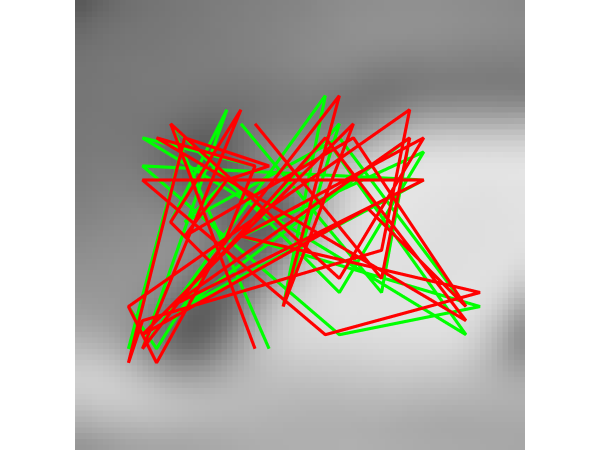}
				\includegraphics[width=0.15\textwidth]{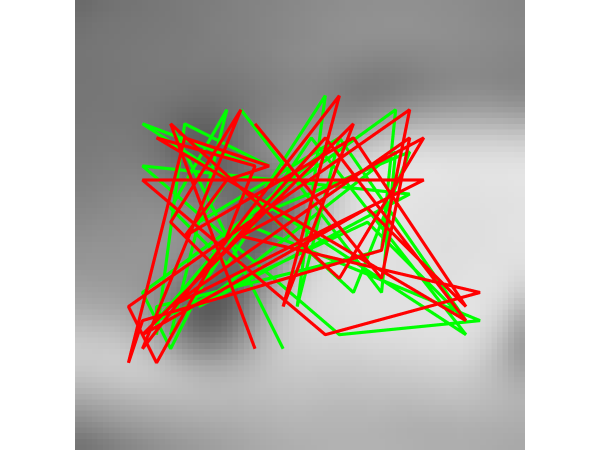}
				\includegraphics[width=0.15\textwidth]{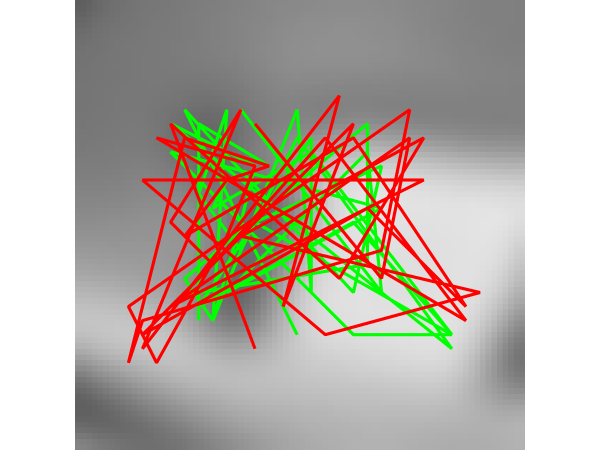}
				\includegraphics[width=0.15\textwidth]{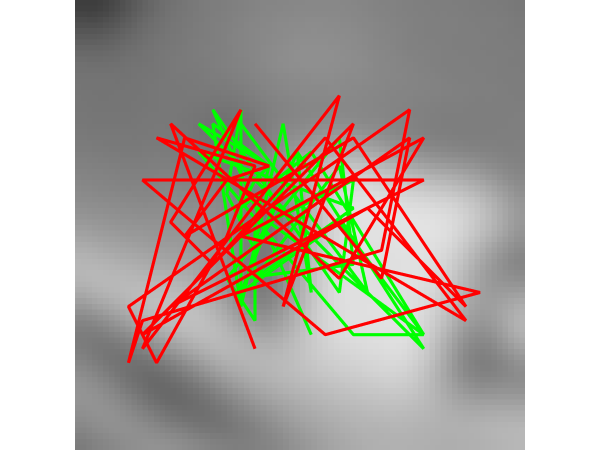}
				\includegraphics[width=0.15\textwidth]{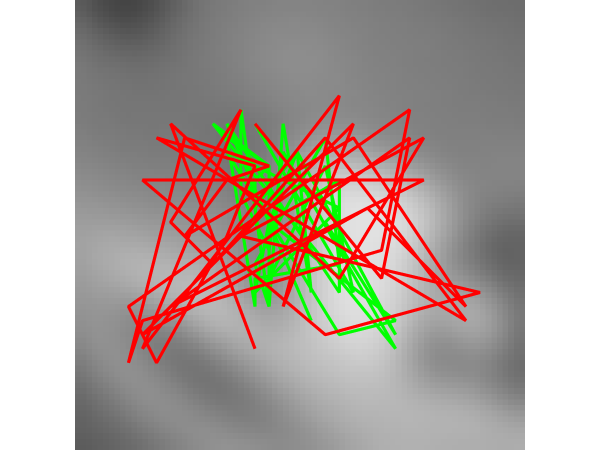}} 
		}	
		\caption{Effect on the test set $\mathbf{Q}$. Red: original tests $\mathbf{Q}$, Green: distorted tests $\mathbf{Q}_d$  and $\mathbf{Q}_f$ in second and third row respectively.} 
		\label{fig:testsCorrespondingTracking}
	\end{minipage}%
\end{figure*}
\section{Simulations}
\label{ch:features:sec:simulations}
To show the viability, performance and properties of our distorted versions of BRIEF (\Seeeq{eq:classCentralRad} and \Seeeq{eq:Omni}), we perform a simulation and various tests.
We simulate a projection of the first image of the popular Graffiti dataset \cite{Mikolajczyk:pami2005} to a virtual camera.
Therefore, the image is mapped to the x-y plane in 3D.
Then, a virtual camera with known exterior orientation and fixed image size moves along the x-axis and observes the Graffiti plane.
Three versions of this sequence are depicted in \Seefig{fig:simulationXtracking}.
The images in each column originate from the same exterior orientation.
The rows represent different interior orientation parametrization.
For the first row, a standard perspective projection is used.
This simulation is used as the source implementation of our distorted versions, as its outcome is supposed to be predictable and equals the behavior of BRIEF.
The second row mirrors a perspective projection with radial distortion (\Seeeq{eq:classCentralRad}).
We set the distortion coefficient to $\xi = -2^{-6}$ for all experiments.
The third row depicts a virtual fisheye camera.
We use the real calibration data from the fisheye camera used in the experimental section.

To study the transformation of the original test set $\mathbf{Q}$, a keypoint is picked on the original Graffiti image. 
Then, we select a patch of fixed size (32x32 pixel in all experiments), around its projection in the virtual camera (red squares in \Seefig{fig:simulationXtracking}).
\SeeFig{fig:patchesCorrespondingTracking} depicts smoothed versions (Gaussian kernel of width 2) of the extracted patches from the image sequences.
The leftmost column shows the original patch from the Graffiti image.
Obviously, the original patch content gets significantly distorted and compressed except, of course, for the standard perspective projection.
This gives rise to two observations.
First, assuming the detector manages to extract the same keypoint over the sequence, the patch content is highly dependent on the position in the image. 
Second, the application of the standard test set $\mathbf{Q}$ on the patch will probably have a different outcome (and thus descriptor) for each patch of the distorted versions.
In addition, this difference will be bigger the farther we move away from the original patch (original camera pose).
This, of course, is unfavorable, as matching performance will drop significantly as the Hamming distance between the same descriptor grows.

To visualize the second observation, we overlay the original test set $\mathbf{Q}$ and the transformed versions of it on some patches of the sequence.
This is depicted in \Seefig{fig:testsCorrespondingTracking}.
The red lines depict the first 20 tests from the original test set $\mathbf{Q}$.
The green lines depict the first 20 tests from the distorted test sets $\mathbf{Q}_d$ and $\mathbf{Q}_f$ in the middle and last row respectively.

Evidently, the outcome of testing pairwise pixel intensities with the distorted tests will be different.
To analyze the magnitude of this difference, we extract the descriptor for each patch of the sequence using $\mathbf{Q}$ and $\mathbf{Q}_d,\mathbf{Q}_f$ respectively.
Then we compute the Hamming distances between the first and each subsequent descriptor of the sequence. 
In general, the Hamming distance between two binary descriptors $d_i$ and $d_j$ is given by:
\begin{equation}
H(\mathbf{d}_i,\mathbf{d}_j) = \mathbf{d}_i \oplus \mathbf{d}_j
\label{eq:Hamming}
\end{equation}
where $\oplus$ is the logical XOR operator.
Ideally, this distances would be zero, which, however, is unrealistic in practice, due to noise, discretization, illumination and viewpoint changes.
Practically, the Hamming distance between the descriptors of equal patches should be small and constant over the sequence.
In other words, the intra-class distance as well as its variance should be as small as possible.

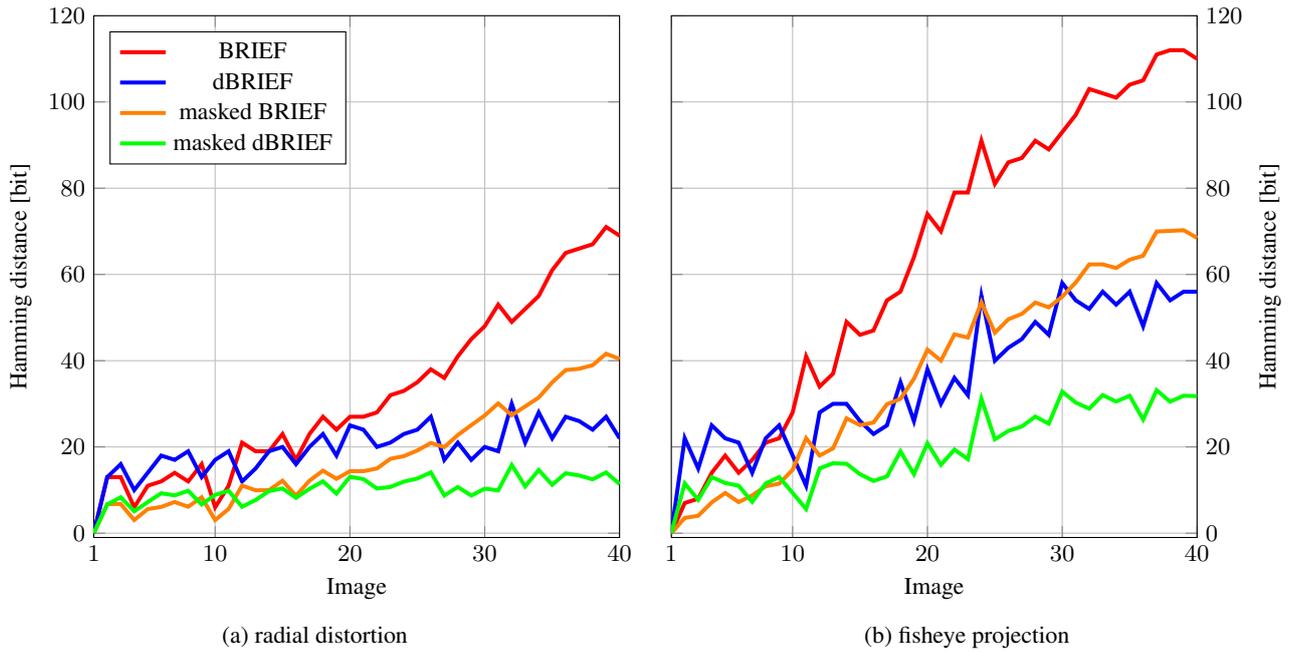
\begin{figure*}
	\subcaptionbox{radial distortion\label{fig:matchErrorRadial}}
	{
		\begin{tikzpicture}[baseline=(current axis.south)]
		\begin{axis}[
		grid=major,
		ymax=120,ymin=-1,
		xmin=1,xmax=40,
		xtick={1,10,20,30,40},
		ylabel={Hamming distance [bit]},
		xlabel={Image},
		yticklabel pos=left,
		width = 0.5\textwidth,
		height = 0.5\textwidth,
		ylabel near ticks,
		xlabel near ticks,
		legend entries={BRIEF, dBRIEF, masked BRIEF, masked dBRIEF},
		legend pos = north west]
		\addplot+[mark=,color=red,line width=1.5pt] 
		table [x index={0},y index={1}]{match_error_off_radial.txt};
		\addplot+[mark=,color=blue,line width=1.5pt]
		table [x index={0},y index={2}]{match_error_off_radial.txt};
		\addplot+[mark=,color=orange,line width=1.5pt]
		table [x index={0},y index={3}]{match_error_off_radial.txt};
		\addplot+[mark=,color=green,line width=1.5pt]
		table [x index={0},y index={4}]{match_error_off_radial.txt};
		\end{axis}
		\end{tikzpicture}
	} \!
	\subcaptionbox{fisheye projection\label{fig:matchErrorFisheye}}
	{
		\begin{tikzpicture}[baseline=(current axis.south)]
		\begin{axis}[
		grid=major,
		ymax=120,ymin=-1,
		xmin=1,xmax=40,
		xtick={1,10,20,30,40},
		ylabel={Hamming distance [bit]},
		xlabel={Image},
		yticklabel pos=right,
		width = 0.5\textwidth,
		height = 0.5\textwidth,
		ylabel near ticks,
		xlabel near ticks]
		\addplot+[mark=,mark size=4,color=red,line width=1.5pt]
		table [x index={0},y index={1}]  {match_error_off_fish.txt};
		\addplot+[mark=,mark size=4,color=blue,line width=1.5pt]
		table [x index={0},y index={2}]  {match_error_off_fish.txt};
		\addplot+[mark=,mark size=4,color=orange,line width=1.5pt]
		table [x index={0},y index={3}]  {match_error_off_fish.txt};
		\addplot+[mark=,mark size=4,color=green,line width=1.5pt]
		table [x index={0},y index={4}]  {match_error_off_fish.txt};
		\end{axis}
		\end{tikzpicture}
	} 
	\caption{Evolution of Hamming distance for the same patch across the simulation sequence but different test sets and masked versions. (a) radial distortion simulation (blue box \SeeFig{fig:simulationXtracking}). (b) fisheye projection simulation (red box \SeeFig{fig:simulationXtracking}).}
	\label{fig:matchErrors}
\end{figure*}

The evolution of the Hamming distance for standard BRIEF ($\mathbf{Q}$) as well as descriptors extracted using $\mathbf{Q}_d$ and $\mathbf{Q}_f$ is depicted in \Seefig{fig:matchErrorRadial} and \Seefig{fig:matchErrorFisheye} respectively.
For now, ignore the two masked version (orange and green graph).
Applying the original test set $\mathbf{Q}$ yields the red graphs (BRIEF), whereas the blue graphs represent the Hamming distance of a distorted version of BRIEF, which we abbreviate with dBRIEF (not to be confused with D-BRIEF \cite{Trzcinski12}).
For the radial distortion version of dBRIEF (\Seefig{fig:matchErrorRadial}), the Hamming distance stays quite constant and is below 30 bits over the whole sequence.
In contrast, the distance between the first and last descriptor of the sequence for the BRIEF descriptor grows continuously and reaches 70 bits which is about a factor of 2.3 larger.
\SeeFig{fig:matchErrorFisheye} shows a similar behavior and BRIEF even reaches 110 bits.
The dBRIEF version for the fisheye camera stays quite constant in the beginning, but then the distance starts to increase around the 20th image and reaches 60 bits.
Still, the error is about two times lower, which is expected to be advantageous during matching.

To test the matching performance of dBRIEF, we first extract 200 keypoints on the original Graffiti image, project their location throughout the image sequences and subsequently extract a descriptor for each patch using BRIEF and dBRIEF respectively.
The patch locations are depicted in \Seefig{fig:orbfeate200sim} on three images of the sequence (10 views in total).
We measure the matching performance using the recognition rate (or recall).
This rate shows the ability of the descriptor to produce correct matches.

Let $\mathbf{D}^i = (\mathbf{d}^i_1,..,\mathbf{d}^i_N)$ be a set of $N$ descriptors extracted in the $i$-th image and $C$ the coincident keypoints detected in both images.
Usually $C=D^i \cap D^j$, but as we do not detect keypoints in every frame but rather project them using the known orientation, $C=D^i=D^j$, i.e. all keypoints appear in all images and we know that each keypoints has a match. 
In the particular simulation $N=200$.
The set of correct matches $S_{true}$ is found by matching each descriptor in the $i$-th image with each descriptor from the $j$-th image.
A match is identified as correct \textit{iff} the distance between two descriptors is minimal.
The recognition rate is computed as follows:
\begin{equation}
\textnormal{recognition rate}(t) = \frac{\# S_{true}}{\# C}
\label{eq:recognitionrate}
\end{equation}
where $\#$ is the count and $t$ a threshold on the descriptor distance.
We match the descriptors of each image of the sequence to the set of descriptors from the first image.
The threshold $t$ for this experiment is not set, i.e. each descriptor will have a corresponding match and $\# C = 200$.

The recognition rates are depicted in \Seefig{fig:recognRateOrbSim}.
Again do not yet consider the masked versions.
\Seefig{fig:recognRateOrbSimRadial} shows the results for the radial distortion version of dBRIEF ($\mathbf{Q}_d$) and \Seefig{fig:recognRateOrbSimFisheye} the fisheye version of dBRIEF ($\mathbf{Q}_f$).
In both cases dBRIEF outperforms BRIEF.
In the radial distortion case, the recognition rate for BRIEF systematically drops from 100\% to about 60\% and in the fisheye case down to 40\%.
The dBRIEF descriptors returns more correct matches and remains above 70\% and around 60\% for radial distortion and fisheye images respectively.
This signifies a relative improvement of 20\% and 50\% respectively.

To get a better understanding of the improvement, we visualize the relative frequencies of Hamming distances of matching and non-matching descriptors in \Seefig{fig:hammingDistanceDistris}.
The blue and red distributions represent the frequencies of matching and non-matching descriptors respectively.
If the distributions do not overlap, only correct matches would be found.
The ultimate goal of improving matching performance for descriptors is to get the distribution overlap as small as possible by decreasing the intra-class distance (make the distributions as tight as possible) and by increasing inter-class distances (distance between the mean of each distribution).

The first two rows of \Seefig{fig:hammingDistanceDistris} visualize the distributions for every third pair of the fisheye sequence.
Mean and variance of the red distribution (non-matching points) remain quite constant. 
The mean of the blue distribution (matching points), however, starts to shift towards the mean of the red distribution. 
In addition the variance increases, i.e. the distribution gets broader.
This can be directly associated with the drop in recognition rate.
To quantify the improvement of dBRIEF over BRIEF, we calculate the Bhattacharyya coefficient $c_{ba}$, which is an approximate measure of the overlap of two distributions.
For the last image pair (1 to 10), the coefficient for BRIEF and dBRIEF is 0.5254 and 0.3947 respectively (27\% lower).

\begin{figure*}
	\centering 
	\includegraphics[width=0.32\textwidth]{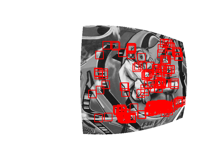}
	\includegraphics[width=0.32\textwidth]{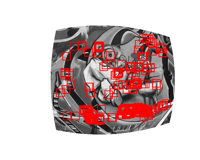}
	\includegraphics[width=0.32\textwidth]{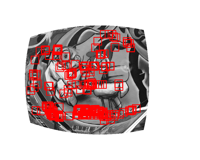}
	\caption{Images with 200 tracked ORB features. Depicted are image 1, 4 and 7 from the radial distortion image set. Fisheye images are not displayed here.}
	\label{fig:orbfeate200sim}
\end{figure*}
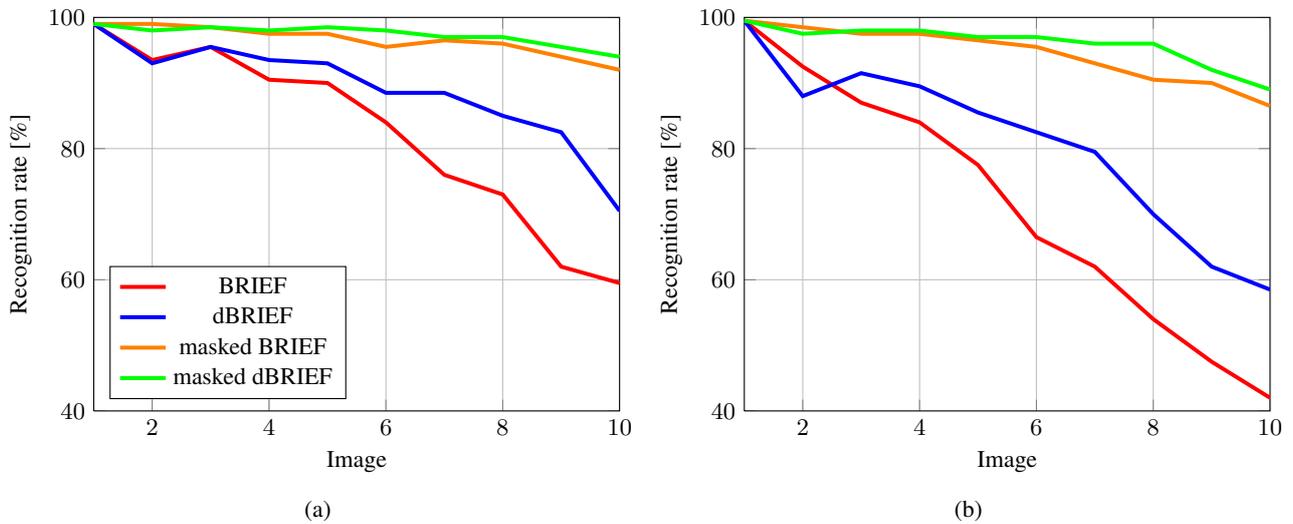
\begin{figure*}
	\subcaptionbox{\label{fig:recognRateOrbSimRadial}}
	{
		\begin{tikzpicture}[baseline=(current axis.south)]
		\begin{axis}[
		grid=major,
		ymax=100,ymin=40,
		xmin=1,xmax=10,
		xtick={2,4,6,8,10},
		ylabel={Recognition rate [\%]},
		xlabel={Image},
		yticklabel pos=left,
		width =  0.5\textwidth,
		height = 0.4\textwidth,
		ylabel near ticks,
		xlabel near ticks,
		legend entries={BRIEF, dBRIEF, masked BRIEF, masked dBRIEF},
		legend columns=1,
		legend pos = south west]
		\addplot+[mark=,color=red,line width=1.5pt]
		table [x index={0},y index={1}]  {recognitonRateOrbSim.txt};
		\addplot+[mark=,color=blue,line width=1.5pt]
		table [x index={0},y index={2}]  {recognitonRateOrbSim.txt};
		\addplot+[mark=,color=orange,line width=1.5pt]
		table [x index={0},y index={3}]  {recognitonRateOrbSim.txt};
		\addplot+[mark=,color=green,line width=1.5pt]
		table [x index={0},y index={4}]  {recognitonRateOrbSim.txt};
		\end{axis}
		\end{tikzpicture}
	}\!
	\subcaptionbox{\label{fig:recognRateOrbSimFisheye}}
	{
		\begin{tikzpicture}[baseline=(current axis.south)]
		\begin{axis}[
		grid=major,
		ymax=100,ymin=40,
		xmin=1,xmax=10,
		xtick={2,4,6,8,10},
		ylabel={Recognition rate [\%]},
		xlabel={Image},
		yticklabel pos=left,
		width = 0.5\textwidth,
		height = 0.4\textwidth,
		ylabel near ticks,
		xlabel near ticks]
		\addplot+[mark=,color=red,line width=1.5pt]
		table [x index={0},y index={1}]  {recognitonRateOrbSimFish.txt};
		\addplot+[mark=,color=blue,line width=1.5pt]
		table [x index={0},y index={2}]  {recognitonRateOrbSimFish.txt};
		\addplot+[mark=,color=orange,line width=1.5pt]
		table [x index={0},y index={3}]  {recognitonRateOrbSimFish.txt};
		\addplot+[mark=,color=green,line width=1.5pt]
		table [x index={0},y index={4}]  {recognitonRateOrbSimFish.txt};
		\end{axis}
		\end{tikzpicture}	
	}
	\caption{Recognition rates for 200 ORB features. Masks are learned for all patches. Hamming distance is computed with two masks. (a) For radial distortion. (b) For the fisheye projection. }
	\label{fig:recognRateOrbSim}
\end{figure*}
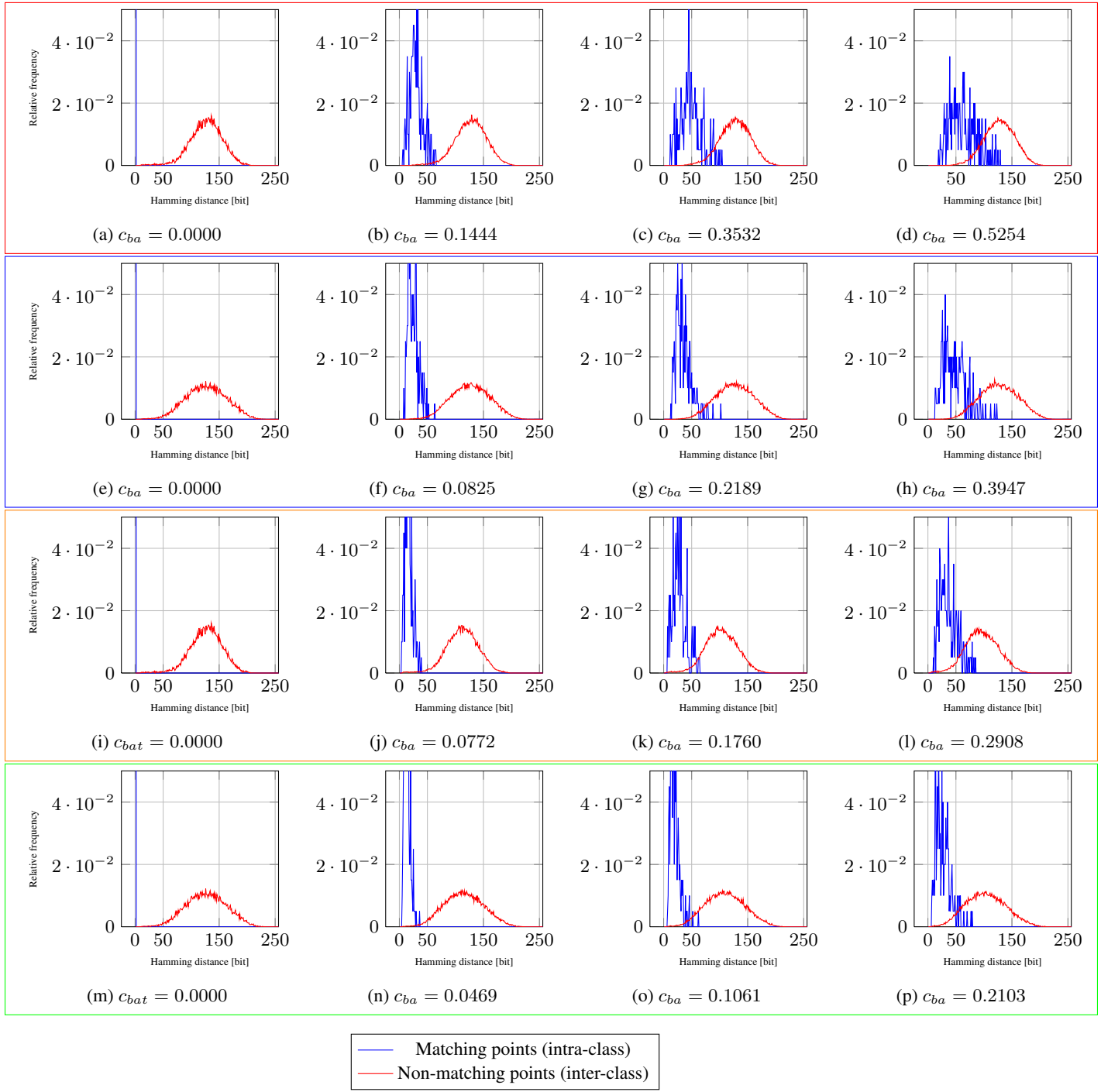
\begin{figure*}
	\centering
	\fcolorbox{red}{white}{
		\subcaptionbox{$c_{ba}=0.0000$}
		{
			\begin{tikzpicture}[baseline=(current axis.south)]
			\begin{axis}[
			grid=major,
			ymax=0.05,ymin=0,xmax=256,
			xtick={0,50,150,250},
			ylabel={\tiny Relative frequency},
			xlabel={\tiny Hamming distance [bit]},
			yticklabel pos=left,
			width = 0.25\textwidth,
			height = 0.25\textwidth,
			ylabel near ticks,
			xlabel near ticks,
			legend entries={Matching points (intra-class), Non-matching points (inter-class)},
			legend to name=legendMatchNonMatch]
			\addplot+[mark=,color=blue]
			table [x index={0},y index={1}]{matchingDistributions1.txt};
			\addplot+[mark=,color=red]
			table [x index={0},y index={2}]{matchingDistributions1.txt};
			\end{axis}
			\end{tikzpicture}
		}
		\subcaptionbox{$c_{ba}=0.1444$}
		{
			\begin{tikzpicture}[baseline=(current axis.south)]
			\begin{axis}[
			grid=major,
			ymax=0.05,ymin=0,xmax=256,
			xtick={0,50,150,250},
			xlabel={\tiny Hamming distance [bit]},
			yticklabel pos=left,
			width = 0.25\textwidth,
			height = 0.25\textwidth,
			ylabel near ticks,
			xlabel near ticks,
			legend pos = north west]
			\addplot+[mark=,color=blue]
			table [x index={0},y index={1}]{matchingDistributions4.txt};
			\addplot+[mark=,color=red]
			table [x index={0},y index={2}]{matchingDistributions4.txt};
			\end{axis}
			\end{tikzpicture}
		}
		\subcaptionbox{$c_{ba}=0.3532$}
		{
			\begin{tikzpicture}[baseline=(current axis.south)]
			\begin{axis}[
			grid=major,
			ymax=0.05,ymin=0,xmax=256,
			xtick={0,50,150,250},
			xlabel={\tiny Hamming distance [bit]},
			yticklabel pos=left,
			width = 0.25\textwidth,
			height = 0.25\textwidth,
			ylabel near ticks,
			xlabel near ticks,
			legend pos = north west]
			\addplot+[mark=,color=blue]
			table [x index={0},y index={1}]{matchingDistributions7.txt};
			\addplot+[mark=,color=red]
			table [x index={0},y index={2}]{matchingDistributions7.txt};
			\end{axis}
			\end{tikzpicture}	
		}
		\subcaptionbox{$c_{ba}=0.5254$}
		{
			\begin{tikzpicture}[baseline=(current axis.south)]
			\begin{axis}[
			grid=major,
			ymax=0.05,ymin=0,xmax=256,
			xtick={50,150,250},
			xlabel={\tiny Hamming distance [bit]},
			yticklabel pos=left,
			width = 0.25\textwidth,
			height = 0.25\textwidth,
			ylabel near ticks,
			xlabel near ticks,
			legend pos = north west]
			\addplot+[mark=,color=blue]
			table [x index={0},y index={1}]{matchingDistributions10.txt};
			\addplot+[mark=,color=red]
			table [x index={0},y index={2}]{matchingDistributions10.txt};
			\end{axis}
			\end{tikzpicture}
		}} \hfill
		\fcolorbox{blue}{white}{
			\subcaptionbox{$c_{ba}=0.0000$}
			{
				\begin{tikzpicture}[baseline=(current axis.south)]
				\begin{axis}[
				grid=major,
				ymax=0.05,ymin=0,xmax=256,
				xtick={0,50,150,250},
				ylabel={\tiny Relative frequency},
				xlabel={\tiny Hamming distance [bit]},
				yticklabel pos=left,
				width = 0.25\textwidth,
				height = 0.25\textwidth,
				ylabel near ticks,
				xlabel near ticks,
				legend pos = north west]
				\addplot+[mark=,color=blue]
				table [x index={0},y index={1}]{matchingDistributionsDist1.txt};
				\addplot+[mark=,color=red]
				table [x index={0},y index={2}]{matchingDistributionsDist1.txt};
				\end{axis}
				\end{tikzpicture}	
			}
			\subcaptionbox{$c_{ba}=0.0825$}
			{
				\begin{tikzpicture}[baseline=(current axis.south)]
				\begin{axis}[
				grid=major,
				ymax=0.05,ymin=0,xmax=256,
				xtick={0,50,150,250},
				xlabel={\tiny Hamming distance [bit]},
				yticklabel pos=left,
				width = 0.25\textwidth,
				height = 0.25\textwidth,
				ylabel near ticks,
				xlabel near ticks,
				legend pos = north west]
				\addplot+[mark=,color=blue]
				table [x index={0},y index={1}]{matchingDistributionsDist4.txt};
				\addplot+[mark=,color=red]
				table [x index={0},y index={2}]{matchingDistributionsDist4.txt};
				\end{axis}
				\end{tikzpicture}		
			}
			\subcaptionbox{$c_{ba}=0.2189$}
			{
				\begin{tikzpicture}[baseline=(current axis.south)]
				\begin{axis}[
				grid=major,
				ymax=0.05,ymin=0,xmax=256,
				xtick={0,50,150,250},
				xlabel={\tiny Hamming distance [bit]},
				yticklabel pos=left,
				width = 0.25\textwidth,
				height = 0.25\textwidth,
				ylabel near ticks,
				xlabel near ticks,
				legend pos = north west]
				\addplot+[mark=,color=blue]
				table [x index={0},y index={1}]{matchingDistributionsDist7.txt};
				\addplot+[mark=,color=red]
				table [x index={0},y index={2}]{matchingDistributionsDist7.txt};
				\end{axis}
				\end{tikzpicture}	
			}
			\subcaptionbox{$c_{ba}=0.3947$}
			{
				\begin{tikzpicture}[baseline=(current axis.south)]
				\begin{axis}[
				grid=major,
				ymax=0.05,ymin=0,xmax=256,
				xtick={0,50,150,250},
				xlabel={\tiny Hamming distance [bit]},
				yticklabel pos=left,
				width = 0.25\textwidth,
				height = 0.25\textwidth,
				ylabel near ticks,
				xlabel near ticks,
				legend pos = north west]
				\addplot+[mark=,color=blue]
				table [x index={0},y index={1}]{matchingDistributionsDist10.txt};
				\addplot+[mark=,color=red]
				table [x index={0},y index={2}]{matchingDistributionsDist10.txt};
				\end{axis}
				\end{tikzpicture}
			}}\hfill
			\fcolorbox{orange}{white}{
				\subcaptionbox{$c_{bat} = 0.0000$}
				{
					\begin{tikzpicture}[baseline=(current axis.south)]
					\begin{axis}[
					grid=major,			
					ymax=0.05,ymin=0,xmax=256,
					xtick={0,50,150,250},
					ylabel={\tiny Relative frequency},
					xlabel={\tiny Hamming distance [bit]},
					yticklabel pos=left,
					width = 0.25\textwidth,
					height = 0.25\textwidth,
					ylabel near ticks,
					xlabel near ticks,
					legend pos = north west]
					\addplot+[mark=,color=blue]
					table [x index={0},y index={1}]{matchingDistributionsMask1.txt};
					\addplot+[mark=,color=red]
					table [x index={0},y index={2}]{matchingDistributionsMask1.txt};
					\end{axis}
					\end{tikzpicture}
				}
				\subcaptionbox{$c_{ba}=0.0772$}
				{
					\begin{tikzpicture}[baseline=(current axis.south)]
					\begin{axis}[
					grid=major,
					ymax=0.05,ymin=0,xmax=256,
					xtick={0,50,150,250},
					xlabel={\tiny Hamming distance [bit]},
					yticklabel pos=left,
					width = 0.25\textwidth,
					height = 0.25\textwidth,
					ylabel near ticks,
					xlabel near ticks,
					legend pos = north west]
					\addplot+[mark=,color=blue]
					table [x index={0},y index={1}]{matchingDistributionsMask4.txt};
					\addplot+[mark=,color=red]
					table [x index={0},y index={2}]{matchingDistributionsMask4.txt};
					\end{axis}
					\end{tikzpicture}		
				}
				\subcaptionbox{$c_{ba}=0.1760$}
				{
					\begin{tikzpicture}[baseline=(current axis.south)]
					\begin{axis}[
					grid=major,
					ymax=0.05,ymin=0,xmax=256,
					xtick={0,50,150,250},
					xlabel={\tiny Hamming distance [bit]},
					yticklabel pos=left,
					width = 0.25\textwidth,
					height = 0.25\textwidth,
					ylabel near ticks,
					xlabel near ticks,
					legend pos = north west]
					\addplot+[mark=,color=blue]
					table [x index={0},y index={1}]{matchingDistributionsMask7.txt};
					\addplot+[mark=,color=red]
					table [x index={0},y index={2}]{matchingDistributionsMask7.txt};
					\end{axis}
					\end{tikzpicture}
				}
				\subcaptionbox{$c_{ba}=0.2908$}
				{
					\begin{tikzpicture}[baseline=(current axis.south)]
					\begin{axis}[
					grid=major,
					ymax=0.05,ymin=0,xmax=256,
					xtick={0,50,150,250},
					xlabel={\tiny Hamming distance [bit]},
					yticklabel pos=left,
					width = 0.25\textwidth,
					height = 0.25\textwidth,
					ylabel near ticks,
					xlabel near ticks,
					legend pos = north west]
					\addplot+[mark=,color=blue]
					table [x index={0},y index={1}]{matchingDistributionsMask10.txt};
					\addplot+[mark=,color=red]
					table [x index={0},y index={2}]{matchingDistributionsMask10.txt};
					\end{axis}
					\end{tikzpicture}		
				}}\hfill
				\fcolorbox{green}{white}{
					\subcaptionbox{$c_{bat} = 0.0000$}
					{
						\begin{tikzpicture}[baseline=(current axis.south)]
						\begin{axis}[
						grid=major,
						ymax=0.05,ymin=0,xmax=256,
						xtick={0,50,150,250},
						ylabel={\tiny Relative frequency},
						xlabel={\tiny Hamming distance [bit]},
						yticklabel pos=left,
						width = 0.25\textwidth,
						height = 0.25\textwidth,
						ylabel near ticks,
						xlabel near ticks,
						legend pos = north west]
						\addplot+[mark=,color=blue]
						table [x index={0},y index={1}]{matchingDistributionsDistMask1.txt};
						\addplot+[mark=,color=red]
						table [x index={0},y index={2}]{matchingDistributionsDistMask1.txt};
						\end{axis}
						\end{tikzpicture}
					}
					\subcaptionbox{$c_{ba}=0.0469$}
					{
						\begin{tikzpicture}[baseline=(current axis.south)]
						\begin{axis}[
						grid=major,
						ymax=0.05,ymin=0,xmax=256,
						xtick={0,50,150,250},
						xlabel={\tiny Hamming distance [bit]},
						yticklabel pos=left,
						width = 0.25\textwidth,
						height = 0.25\textwidth,
						ylabel near ticks,
						xlabel near ticks,
						legend pos = north west]
						\addplot+[mark=,color=blue]
						table [x index={0},y index={1}]{matchingDistributionsDistMask4.txt};
						\addplot+[mark=,color=red]
						table [x index={0},y index={2}]{matchingDistributionsDistMask4.txt};
						\end{axis}
						\end{tikzpicture}
					}
					\subcaptionbox{$c_{ba}=0.1061$}
					{
						\begin{tikzpicture}[baseline=(current axis.south)]
						\begin{axis}[
						grid=major,
						ymax=0.05,ymin=0,xmax=256,
						xtick={0,50,150,250},
						xlabel={\tiny Hamming distance [bit]},
						yticklabel pos=left,
						width = 0.25\textwidth,
						height = 0.25\textwidth,
						ylabel near ticks,
						xlabel near ticks,
						legend pos = north west]
						\addplot+[mark=,color=blue]
						table [x index={0},y index={1}]{matchingDistributionsDistMask7.txt};
						\addplot+[mark=,color=red]
						table [x index={0},y index={2}]{matchingDistributionsDistMask7.txt};
						\end{axis}
						\end{tikzpicture}
					}
					\subcaptionbox{$c_{ba}=0.2103$}
					{
						\begin{tikzpicture}[baseline=(current axis.south)]
						\begin{axis}[
						grid=major,
						ymax=0.05,ymin=0,xmax=256,
						xtick={0,50,150,250},
						xlabel={\tiny Hamming distance [bit]},
						yticklabel pos=left,
						width = 0.25\textwidth,
						height = 0.25\textwidth,
						ylabel near ticks,
						xlabel near ticks,
						legend pos = north west]
						\addplot+[mark=,color=blue]
						table [x index={0},y index={1}]{matchingDistributionsDistMask10.txt};
						\addplot+[mark=,color=red]
						table [x index={0},y index={2}]{matchingDistributionsDistMask10.txt};
						\end{axis}
						\end{tikzpicture}
					}}		
					\begin{center}
						\ref{legendMatchNonMatch}
					\end{center}		
					\caption{Hamming distance distributions for matching (intra-class) and non-matching (inter-class) binary descriptors for the fisheye camera simulation. Each column represents the distributions for the image pairs 1-1,1-3,1-7 and 1-10 respectively. $c_{ba}$ is the Bhattacharyya coefficient for the corresponding plot. First row (a)-(d) BRIEF, second row (e)-(h) dBRIEF ($\mathbf{S}_f$), third row (i)-(l) mBRIEF, fourth row (m)-(p) mdBRIEF. }
					\label{fig:hammingDistanceDistris}
				\end{figure*}

\section{Improving dBRIEF by On-line Mask Learning}
\label{ch:features:sec:mdBRIEF}
As mentioned frequently throughout this paper, one way of improving the matching performance is to decrease the intra-class variance whilst simultaneously increasing inter-class distance between matching and non-matching descriptors.
Most learning methods \cite{trzcinski2013boosting,strecha2012ldahash,trzcinski2012learning,Trzcinski12,han2015matchnet,simo2015discriminative,lategahn2013learn} optimize both criteria offline and simultaneously for large datasets of corresponding patches.
To minimize the intra-class variance, patches of the same keypoint are needed.
Even though the datasets provide multiple instances of the same patch, the optimization of the intra-class variance is performed over descriptors that originate from the same image category and contain similar image characteristics and patch content.
As pointed out by \cite{balntas2015bold}, an on-line adaption of the descriptor creation, i.e. a per patch optimization of the intra-class variance for a specific problem/camera/scene should be beneficial and increase matching performance.

This is, however, computationally too expensive for sophisticated, yet cumbersome optimization algorithms that learn functions globally that map the patch content to descriptors.
Thus, we will follow the ideas of \cite{rublee2011orb,balntas2015bold,richardsontailoredbrief} and adapt the binary descriptors online.
Fortunately, this is not too expensive to compute for binary descriptors and exploits the properties of binary tests, briefly resumed in the following.
\subsection{Properties of Binary Tests}
In general, the outcome of a binary test is described by a Bernoulli distribution, given the probability $\alpha$ of success (value 1) and $\beta=1-\alpha$ failure (value 0).
The mean of a binary test is given by $\alpha$ and the variance by $\sigma^2 = \alpha(1-\alpha)$.

\begin{figure*}
	\centering
	\subcaptionbox{Inter-class optimization	\label{fig:globalDescOptim}}
	{\includegraphics[clip=true, trim = 1mm 88mm 260mm 1mm, width=0.3\textwidth]{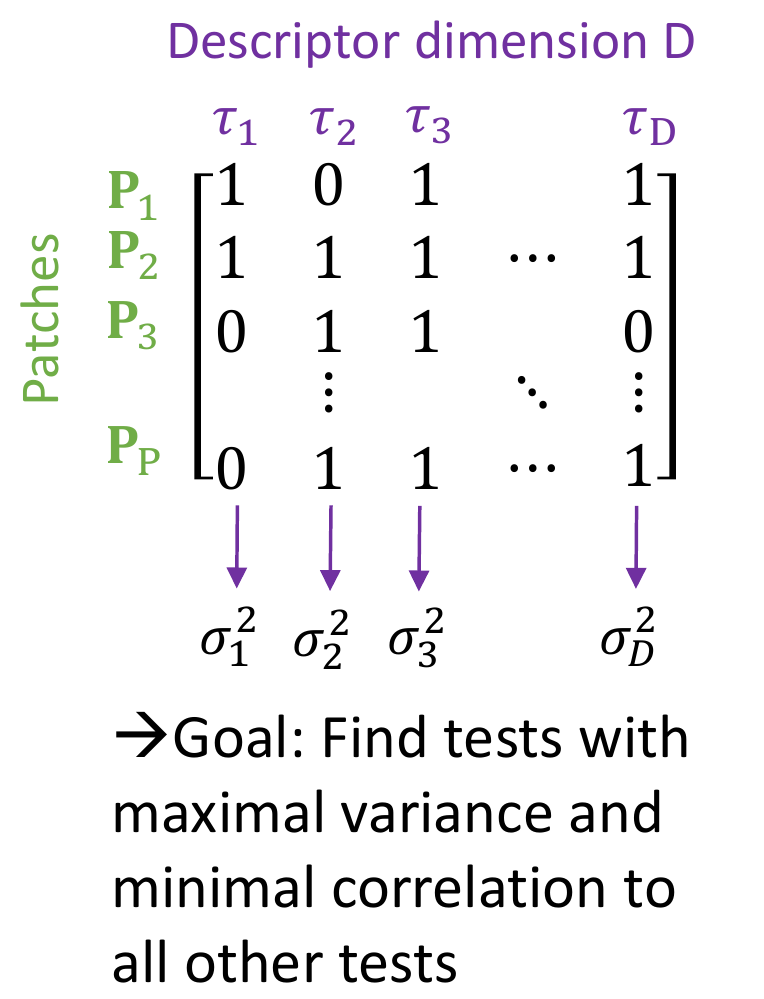}}
	\subcaptionbox{Intra-class optimization	\label{fig:LocalDescOptim}}
	{\includegraphics[clip=true, trim = 1mm 88mm 260mm 1mm,width=0.3\textwidth]{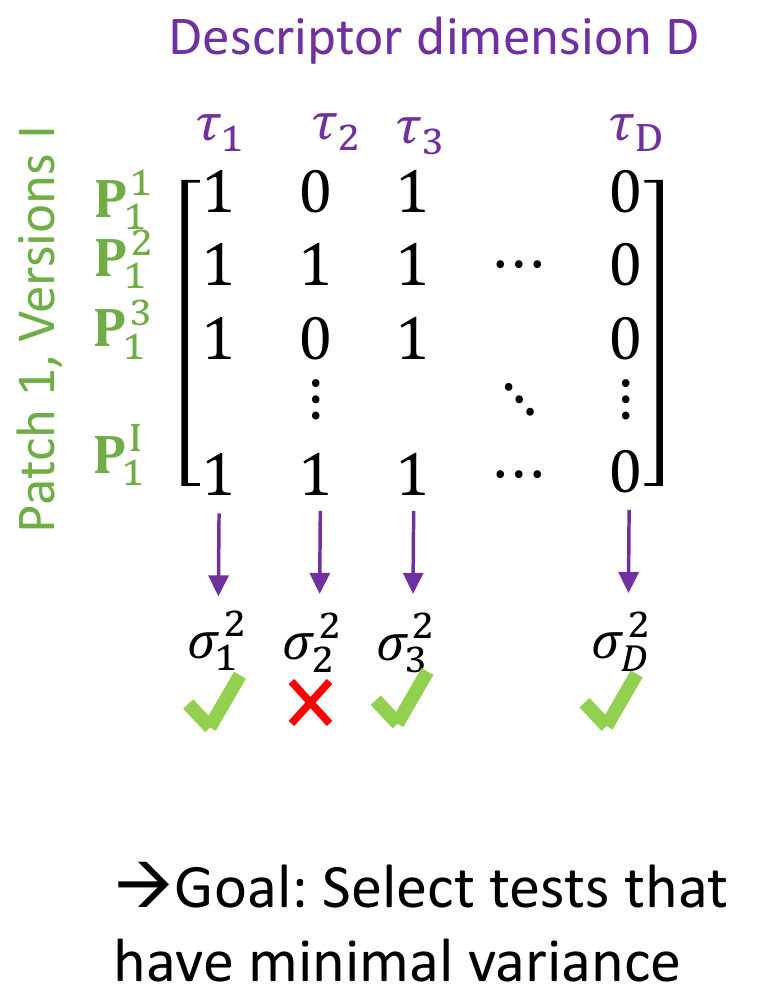}}
	\caption{(a) The matrix has dimension $P \times D$. The goal is to increase the inter-class distances. Each column represents one test $\tau$ of the descriptor and each row represents the results of applying $\tau$ to patch $\mathbf{P}$. Thus a variance $\sigma^2$ of one particular test over all patches can be computed. (b) The matrix has dimension $I \times D$. To minimize intra-class distances tests that have zero variance across different versions of the same patch are selected.}
	\label{fig:LDAoptimization}
\end{figure*}

Now, let $\mathbf{A}$ be a $P \times D$ matrix of $P$ binary descriptors of dimension $D$.
This matrix is depicted in \Seefig{fig:globalDescOptim}.
Each column of $\mathbf{A}$ represents one binary test (descriptor dimension).
The outcome of this binary test $\tau_d$ on a particular patch $\mathbf{P}$ are stored in the rows of $\mathbf{A}$ at position $a_{p,d}$.
To learn a globally best set of binary tests, the variance $\sigma^2_d=\alpha_d(1-\alpha_d)$ of a particular test $\tau_d$ should be high \cite{rublee2011orb}, where $\alpha_d = \sum_{P}a_{p,d} / P$ is the ratio of ones for test $\tau_d$ on a patch $\mathbf{P}$. 
This is reasonable, as the outcome should be as variable as possible on different patches (represented by the rows of $\mathbf{A}$).

To learn a locally best set of binary tests, the rows are replaced by different instances/representations of the same patch.
This can either be achieved by warping the patch using small rotations or affine deformations or by applying the deformations to the test set.
Then, the goal is to find those tests, that have the lowest variance (preferably zero), over all instances of the patch.
This is depicted in \Seefig{fig:LocalDescOptim}.
In this example, test two has non-zero variance and is suppressed.

In the following two sections, we briefly describe the off-line and on-line learning for dBRIEF and subsequently present results on simulated and real data.

%

\subsection{Off-line Learning}
\label{ch:features:sec:offline}
We follow the unsupervised learning of a set of high variance tests with low correlation which was first presented in \cite{rublee2011orb}.
Since then, various authors \cite{balntas2015bold,levi2015latch,alahi2012freak} adopted the learning scheme.
Very recently, the authors of BinBoost \cite{trzcinski2013boosting} proposed a new learning method for intensity test configurations \cite{Trzcinski15} by generalization of the employed boosting method for gradient based descriptors.
Although they achieve a performance gain over the greedy approach of \cite{rublee2011orb}, we use latter to learn our test set off-line, and postpone a comparison to \cite{Trzcinski15} to future work.

Given a patch of size $S \times S$, the number of possible intensity comparisons within this patch is $D=({S^2 \atop 2})$.
Such a binary pixel intensity test is defined as in \Seeeq{eq:testBrief} and we construct a set $\mathbf{Q}_{all}$ of all possible test configurations.
For a patch size of $S=32$, that is used for all experiments, the total number of all tests would be 523776.
To reduce this number, we omit tests along the patch border and tests that have a distance $d < 3$ or $d > 9S/10$, leaving us with a total number of $377650$ possible binary tests.
Then, we extract 20000 patches from two different datasets.
The first dataset consists of images from the PASCAL visual object classes challenge \cite{everingham2006pascal} and is the same as used in \cite{rublee2011orb}.
The second is a fisheye image dataset.
To obtain various types of fisheye images, we performed a Internet search for fisheye images and randomly downloaded 300 images.

In addition the extracted patches are computed from keypoints obtained by three different detectors.
We chose ORB, AKAZE and SURF to detect keypoints on both datasets.
All three descriptors use different methods to detect a keypoint and compute its dominant orientation.
The loss in test variance in the original BRIEF descriptor comes basically from rotating the test set according to the keypoint orientation \cite{rublee2011orb}.
During learning, each test is rotated before applying it to the patch, thus the learned results will depend on the estimated rotation.
This study is motivated by the fact that none of the papers that adopted this greedy learning scheme investigated the impact of using different images and detectors on the overall performance of the final descriptor.
The fact that the BOLD descriptor is learned on a different dataset than ORB, but has a better performance (without masks) indicates, that this influences the result.

The off-line learning is conducted in two steps.
First, each test is performed on each patch.
In theory, this results in a matrix $\mathbf{A}$ of size $P \times D$.
As this matrix would not fit into memory, we directly compute the variance $\sigma^2_d$ for one test $\tau_d$ applied to all patches.
Then, we sort the $D$ binary tests according to their variance and save the sorted test set to $\mathbf{Q}_{sorted} = (\tau^s_1,..,\tau^s_D)$.

In the second step, we try to identify tests that have a correlation $c < t_c$, where $t_c$ is a threshold.
In order to find such low correlation tests, we start with the first test from $\mathbf{Q}_{sorted}$ and save it to the final test set $\mathbf{Q} = (\tau_1,..,\tau_D)$.
Then, we iterate through all patches, apply one sorted test $\tau^s_d$ and compare the correlation to all tests from the final test set $\mathbf{Q}$.
The correlation is computed as \cite{balntas2015bold}:
\begin{equation}
c = \mathopen| \frac{2}{P} \sum_{p=1}^{P} \mathopen| \tau_{d,p}-\tau^s_{d,p}\mathclose| - 1 \mathclose|
\label{eq:correlation}
\end{equation}
This value thus varies between 0 and 1 where 0 is no and 1 perfect correlation, respectively.
We start with a correlation threshold of $t_c = 0.2$ and increase this value by $0.1$ every time we iterated over all tests $\mathbf{Q}_{sorted}$.
This is conducted until a desired number of tests is found (e.g. 512) that yield the final descriptor.
\subsection{On-line Mask Learning}
\label{ch:features:sec:online}
After having identified a high variance-low correlation test set that maximizes inter-class distance, we can proceed by trying to reduce the intra-class variance per patch. The theoretical background is identical to the one proposed in \cite{balntas2015bold}.
Our contribution is to learn masks for our distorted descriptor dBRIEF.

Instead of learning a new test set $\mathbf{Q}$ for each patch (which would be infeasible in practice), we learn a mask $\mathbf{L}=(l_1,...,l_d)$ of length $D$, with $l_d \in \{0,1\}$.
This mask suppresses individual tests during matching if $l_d=0$, i.e. the result of a particular test $d$ has no influence on the Hamming distance between two descriptors.

To learn such a mask online, we find the tests that are stable under various transformations of the original patch.
In other words, we try to find tests that have zero intra-class variance.
Instead of re-sampling the patch, the off-line learned test set $\mathbf{Q}$ is transformed.
Those transformations have to be carried out online, with every extraction of a new keypoint.
Luckily, as pointed out by \cite{balntas2015bold,cai2011learning} and confirmed by own experiments, two small random rotations work better than performing many full affine transformations. 

Now, if a keypoint is detect, we anchor the test set $\mathbf{Q}$ to the undistorted location of the keypoint using \Seeeq{eq:undistTesSet}.
Here, we perform two small rotations of the test set and project the three test sets back to the image yielding $\mathbf{Q}^{0,1,2}_d$ or $\mathbf{Q}^{0,1,2}_f$ respectively, where the superscript 0 indicates the original test set and 1,2 are the rotated versions.
Subsequently, zero variance tests are identified by calculating the variance between each test of $\mathbf{Q}^{0}$ and $\mathbf{Q}^{1,2}$.
This is schematically depicted in \Seefig{fig:LocalDescOptim}.

During matching a masked Hamming distance has to be computed.
Given two descriptors $d_i$ and $d_j$ with corresponding learned masks $l_i$ and $l_j$ the distance is computed as:
\begin{equation}
H_{masked}(d_i,d_j,m_i,m_j) = \frac{1}{o_i}l_i \wedge d_i \oplus d_j + \frac{1}{o_j}l_j \wedge d_i \oplus d_j
\label{eq:maskedHamming}
\end{equation}
where $o_i,o_j$ represent the total number of ones in the masks respectively and $\oplus$ is the logical XOR operator.
The masking is achieved by the logical AND $\wedge$ operator that suppresses the contribution of a particular test to the Hamming distance.

To compare the performance improvements over the standard BRIEF and dBRIEF descriptor, we analyze the masked version with the same simulation used for the dBRIEF descriptor.
The Hamming distance error for one patch of the simulation over the entire sequence is depicted in \Seefig{fig:matchErrors}.
The orange and green graphs depict the masked version of BRIEF and dBRIEF respectively.
Clearly, masking unstable tests for each patch decreases the Hamming distance between the descriptors of the same patch, and thus the intra-class variance.
We repeat the experiment for 200 keypoints and calculate the recognition rate.
The result is depicted in \Seefig{fig:recognRateOrbSim}.
The masked versions outperform BRIEF and dBRIEF w.r.t the recognition rate.
This effect can be further investigated by plotting the Hamming distance distributions for matching and non-matching descriptors and calculating the Bhattacharyya coefficient between them.
This is depicted in the last two rows of \Seefig{fig:hammingDistanceDistris}.
The overlap between the distributions for image pair 1-10 for dBRIEF and masked dBRIEF (last plot in red and green box) is $c_{ba}=0.3947$ and $c_{ba}=0.2103$ respectively and thus the overlap of dBRIEF is about 85\% higher.
In general, the intra-class distributions are a lot tighter for the masked versions, showing the effect of reducing the intra-class variance, by masking non-zero variance tests.
In the remainder of this work, the masked version of \textbf{dBRIEF} is called \textbf{mdBRIEF}.
	
%
%
%
%
\section{Results on Real Data}
\label{ch:features:sec:results}
In this Section, the performance of dBRIEF and mdBRIEF is evaluated against the state-of-the-art on a real data sets.
The results for BOLD are based on the code provided by the authors.
For all other descriptors and detectors we use the standard implementations of all methods that are part of OpenCV\footnote[1]{All results in this thesis correspond to OpenCV version 3.1. which was the most up to date version at the time of writing}.

To evaluate the matching performance, we compute the recognition rate and '1-precision' curves according to \cite{Mikolajczyk:pami2005,lourencco2012srd}.
The recognition rate (recall) was already defined in  \SeeSec{ch:features:sec:simulations} and measures the descriptors capability of identifying matches from the total set of available points.
This measure is complemented by the correctness of those detected correspondences and is expressed in terms of the precision.
Let $\mathbf{D}^i,\mathbf{D}^j$ be a set of descriptors extracted from patches around keypoints detected in image $i$ and $j$ respectively. 
Then the coincident keypoints, i.e. points that are inside the image border and detected in both images are found by projecting the points from image $i$ to image $j$ using a given ground truth transformation (e.g. homography for planar scenes) and taking the nearest neighbor in a radius of 3 pixels.
This yields the set of ground truth matching points $C = \mathbf{D}^i \cap \mathbf{D}^j$.

Then we perform Brute Force matching of descriptors, with an increasing Hamming distance threshold $t$, i.e. a match is identified as correct \textit{iff} the distance between two descriptors is minimal and below $t$.
The set of all matches identified by the algorithm is then given by $V$.
From this set the correct $S_{true}$ and incorrect matches can be identified using the ground truth set $C$.
The precision and recognition rate are then given by:
\begin{equation}
\textnormal{recognition rate}(t) = \frac{\# S_{true}}{\# C}, \quad \textnormal{1-precision}(t) = 1-\frac{\# S_{true}}{V}
\label{eq:precision}
\end{equation}
Increasing the threshold $t$ and computing both measures yields a so-called PR-curve.
Assuming a perfect descriptor this curve would pass the point (0,1), i.e. a precision and recognition rate of 1.
In practice, the PR-curve of the best performing descriptor is the one for which the distance to this point is minimal.
If there is more than one image pair, we calculate the mean over the PR-curves for each image pair.
\begin{figure*}
	\includegraphics[width=\textwidth]{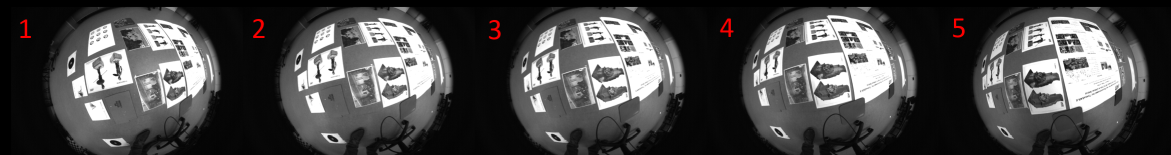}
	\caption{Five of the 11 fisheye images from test case 1 used for descriptor evaluation. Subsequent images are related by a homography (planar scene). The images were recorded with the back looking camera of the helmet MCS (\SeeSec{ch:grundlagen:sec:data}).}
	\label{fig:homographySequence}
\end{figure*}
\paragraph{Fisheye Sequence}
In a first test, we simulate the tracking of subsequent camera images, by recording $I=11$ images $I_i$ from a planar scene with a fisheye camera.
\SeeFig{fig:homographySequence} depicts five images of this sequence.
To assess the matching performance of our proposed method, ground truth matches between the frames have to be determined.
As the scene is planar, the mapping relating subsequent frames is given by a planar projective transformation $\mathbf{H}_{i-1,i}$ from frame $i-1$ to frame $i$.
Thus in a first step, detected image points in the fisheye images have to be undistorted.
This is achieved by using the transformation sequence from \Seeeq{eq:undist}-\SeeEq{eq:undistEbeneOben}.
To compute ground truth homographies between subsequent images, we adopt the method used in \cite{lourencco2012srd}.
We first align the images by manually selecting more than 10 corresponding points in each image pair yielding initial homographies $\mathbf{H}^{init}_{i-1,i}$.
To refine the result, we warp image $i-1$ to image $i$ and extract a large number of AKAZE keypoints with SURF descriptors.
Note, that it is possible to use standard descriptors here, instead of fisheye optimized methods, as the baseline between frames is rather small and the images are undistorted, pre-warped and keypoints are mainly extracted close to the image center. 
Then, putative correspondences between subsequent images of the sequence are found by Brute-Force matching of all SURF descriptors followed by outlier removal using PROSAC \cite{chum2005matching} as implemented in OpenCV.
This yields refined homographies $\mathbf{H}^{fine}_{i-1,i}$ and the final ground truth homographies can be computed by concatenating initial and refined homographies:
$\mathbf{H}^{gt}_{i-1,i} = \mathbf{H}^{fine}_{i-1,i} \mathbf{H}^{init}_{i-1,i}$.

We test different properties of dBRIEF and mdBRIEF on this data set.
First, the impact of learning the fixed test set on different data sets, as well as the descriptor dimension is evaluated.
The results are depicted in \Seefig{fig:descDimAndDetector}.
We extract 200 corners using ORB on 4 octaves over all images.
In this experiment, we do not learn masks, but rotate the tests with the given orientation of the ORB detector.
In \Seefig{fig:DescDimEffect}, it is visible that a larger descriptor dimension leads to better results. 
This effect, however, saturates for a descriptor with more than 256 bits and the performance for 128 bits is already better than a standard 256 bit BRIEF descriptor which uses random tests.
Thus we chose this dimension for dBRIEF and mdBRIEF descriptors\footnote[1]{However, we learned a full test set of 512 tests and thus this number is adjustable. If run-time and storage is not an important issue, performance will certainly be better with more tests.}.

Finally, we benchmark dBRIEF (128bit) and mdBRIEF (128bit) against state-of-the-art binary descriptors, i.e. BRISK (512bit), MLDB (488bit), ORB (256bit) and BOLD (512bit).
The results are depicted in \SeeFig{fig:fishSequStateOfArt}.
In \SeeFig{fig:fishSequOri}, we accounted for the patch orientation estimated by the detector, although the sequence contains almost no rotation about the view direction.
BOLD performs best, followed by mdBRIEF.
Note, however, that the mdBRIEF descriptor is four times smaller than the BOLD descriptor.
The distorted descriptor dBRIEF still performs better than BRISK which also has a length of 512bit.
In \SeeFig{fig:fishSequNoOri}, we omit the patch orientation for BOLD, dBRIEF and mdBRIEF.
Now, all descriptors without orientation perform better and mdBRIEF almost reaches the same performance as BOLD.
  
\begin{figure*}
	\subcaptionbox{Different detectors and datasets used for descriptor learning }
	{
		\begin{tikzpicture}[baseline=(current axis.south)]
		\begin{axis}[
		grid=major,
		ymax=1,ymin=0,
		xmax=0.6,xmin=0,
		xtick={0.2,0.4,0.6,0.8},
		xlabel={1-precision},
		ylabel={Recognition rate (recall)},
		yticklabel pos=left,
		width = 0.5\textwidth,
		height = 0.5\textwidth,
		ylabel near ticks,
		xlabel near ticks,
		legend entries={\tiny ORB-perspective,\tiny ORB-fisheye,\tiny ORB-ocamfisheye, \tiny SURF-perspective, \tiny SURF-fisheye,\tiny AKAZE-perspective,\tiny AKAZE-fisheye, \tiny random tests},
		legend pos = south east]
		\addplot+[mark=+,color=red,line width=0.5pt] 
		table [x index={0},y index={1}]{pr_fisheye_offline_diff_detector.txt};
		\addplot+[mark=triangle,color=red,line width=0.5pt] 
		table [x index={2},y index={3}]{pr_fisheye_offline_diff_detector.txt};
		\addplot+[mark=o,color=red,line width=0.5pt] 
		table [x index={4},y index={5}]{pr_fisheye_offline_diff_detector.txt};
		\addplot+[mark=+,color=green,line width=0.5pt] 
		table [x index={6},y index={7}]{pr_fisheye_offline_diff_detector.txt};
		\addplot+[mark=triangle,color=green,line width=0.5pt] 
		table [x index={8},y index={9}]{pr_fisheye_offline_diff_detector.txt};
		\addplot+[mark=+,color=blue,line width=0.5pt] 
		table [x index={10},y index={11}]{pr_fisheye_offline_diff_detector.txt};
		\addplot+[mark=triangle,color=blue,line width=0.5pt] 
		table [x index={12},y index={13}]{pr_fisheye_offline_diff_detector.txt};
		\addplot+[mark=,color=black,line width=0.5pt] 
		table [x index={14},y index={15}]{pr_fisheye_offline_diff_detector.txt};
		\end{axis}
		\end{tikzpicture}
	} \!
	\subcaptionbox{Performance for different dimensions \label{fig:DescDimEffect}}
	{
		\begin{tikzpicture}[baseline=(current axis.south)]
		\begin{axis}[
		grid=major,
		ymax=1.0,ymin=0,
		xmax=0.6,xmin=0,
		xtick={0.2,0.4,0.6,0.8},
		xlabel={1-precision},
		ylabel={Recognition rate (recall)},
		yticklabel pos=left,
		width = 0.5\textwidth,
		height = 0.5\textwidth,
		ylabel near ticks,
		xlabel near ticks,
		legend entries={\tiny dBRIEF-32,\tiny dBRIEF-64,\tiny dBRIEF-128, \tiny dBRIEF-256,\tiny dBRIEF-512, \tiny BRIEF-256},
		legend pos = south east]
		\addplot+[mark=+,color=blue,line width=1pt]
		table [x index={0},y index={1}]{pr_fisheye_offline_learned_dimension.txt};
		\addplot+[mark=+,color=cyan,line width=1pt]
		table [x index={2},y index={3}]{pr_fisheye_offline_learned_dimension.txt};
		\addplot+[mark=+,color=orange,line width=1pt]
		table [x index={4},y index={5}]{pr_fisheye_offline_learned_dimension.txt};
		\addplot+[mark=+,color=green,line width=1pt]
		table [x index={6},y index={7}]{pr_fisheye_offline_learned_dimension.txt};	
		\addplot+[mark=+,color=magenta,line width=1pt]
		table [x index={8},y index={9}]{pr_fisheye_offline_learned_dimension.txt};
		\addplot+[mark=,color=black,line width=2pt]
		table [x index={10},y index={11}]{pr_fisheye_offline_learned_dimension.txt};
		\end{axis}
		\end{tikzpicture}
	} 
	\caption{Different investigations on offline learned test sets. (a) Depicts the descriptor performance for test sets learned on different detectors and datasets compared to each other and a random test set (BRIEF). The dimension of each test set is set to 256. \textit{ORB-ocamfisheye} was learned on datasets from \SeeSec{ch:grundlagen:sec:data} and tests were distorted during the off-line learning process. (b) The descriptor dimensions is tested. Now dBRIEF uses the best performing off-line learned test set from the left figure. dBRIEF-128 already performs better than BRIEF-256 (also distorted but random tests). The performance gain from 256 to 512 bit is rather small.}
	\label{fig:descDimAndDetector}
\end{figure*}
\begin{figure*}
\subcaptionbox{Including patch orientation \label{fig:fishSequOri}}
{
		\begin{tikzpicture}[baseline=(current axis.south)]
		\begin{axis}[
		grid=major,
		ymax=1.0,ymin=0,
		xmax=0.7,xmin=0,
		xtick={0.2,0.4,0.6,0.8},
		xlabel={1-precision},
		ylabel={Recognition rate (recall)},
		yticklabel pos=left,
		width = 0.5\textwidth,
		height = 0.4\textwidth,
		ylabel near ticks,
		xlabel near ticks,
		legend entries={BRISK-512, AKAZE-MLDB-488, ORB-256, dBRIEF-128, BOLD-512, mdBRIEF-128},
		legend columns = 2,
		legend to name=legendPRCurves1]
		\addplot+[mark=+,color=blue,line width=1pt]
		table [x index={0},y index={1}]{pr_fish_sequence_wrot.txt};
		\addplot+[mark=+,color=black,line width=1pt]
		table [x index={2},y index={3}]{pr_fish_sequence_wrot.txt};
		\addplot+[mark=+,color=orange,line width=1pt]
		table [x index={4},y index={5}]{pr_fish_sequence_wrot.txt};
		\addplot+[mark=+,color=green,line width=1pt]
		table [x index={6},y index={7}]{pr_fish_sequence_wrot.txt};
		\addplot+[mark=+,color=red,line width=1pt]
		table [x index={8},y index={9}]{pr_fish_sequence_wrot.txt};
		\addplot+[mark=+,color=green,line width=1pt]
		table [x index={10},y index={11}]{pr_fish_sequence_wrot.txt};
		\end{axis}
		\end{tikzpicture}
}
\subcaptionbox{Ignoring patch orientation for dBRIEF, mdBRIEF and BOLD  \label{fig:fishSequNoOri}}
{
		\begin{tikzpicture}[baseline=(current axis.south)]
		\begin{axis}[
		grid=major,
		ymax=1.0,ymin=0,
		xmax=0.7,xmin=0,
		xtick={0.2,0.4,0.6,0.8},
		xlabel={1-precision},
		ylabel={Recognition rate (recall)},
		yticklabel pos=left,
		width = 0.5\textwidth,
		height = 0.4\textwidth,
		ylabel near ticks,
		xlabel near ticks]
		\addplot+[mark=+,color=blue,line width=1pt]
		table [x index={0},y index={1}]{pr_fish_sequence_worot.txt};
		\addplot+[mark=+,color=black,line width=1pt]
		table [x index={2},y index={3}]{pr_fish_sequence_worot.txt};
		\addplot+[mark=+,color=orange,line width=1pt]
		table [x index={4},y index={5}]{pr_fish_sequence_worot.txt};
		\addplot+[mark=+,color=green,line width=1pt]
		table [x index={6},y index={7}]{pr_fish_sequence_worot.txt};
		\addplot+[mark=+,color=red,line width=1pt]
		table [x index={8},y index={9}]{pr_fish_sequence_worot.txt};
		\addplot+[mark=+,color=green,line width=1pt]
		table [x index={10},y index={11}]{pr_fish_sequence_worot.txt};		
		\end{axis}
		\end{tikzpicture}
}
\begin{center}
	\ref{legendPRCurves1}
\end{center}	
\caption{Depicted are the precision-recall curves for the fisheye sequence with 10 image pairs and almost no rotation about the camera view axis. Note, that dBRIEF and mdBRIEF use only 128 test, i.e. the descriptor has 128bits. (a) descriptor were extracted including patch orientation. (b) descriptors for BOLD, dBRIEF and mdBRIEF were extracted without rotating the descriptor pattern. }
\label{fig:fishSequStateOfArt}
\end{figure*}

\begin{figure*}
	\centering
	\subcaptionbox{Dataset Firewire. \label{fig:lowDistortionSRD}}
	{
		\begin{tikzpicture}[baseline=(current axis.south)]
		\begin{axis}[
		grid=major,
		ymax=1.0,ymin=0,
		xmax=0.7,xmin=0,
		xtick={0.2,0.4,0.6,0.8},
		xlabel={1-precision},
		ylabel={Recognition rate (recall)},
		yticklabel pos=left,
		width = 0.5\textwidth,
		height = 0.4\textwidth,
		ylabel near ticks,
		xlabel near ticks,
		legend entries={BRISK-512, AKAZE-MLDB-488, ORB-256, mdBRIEF-128, BOLD-512, mdBRIEF-128 w/o rotation, BOLD-512 w/o rotation},
		legend columns = 2,
		legend to name=legendPRCurves]
		\addplot+[mark=+,color=blue,line width=1pt]
		table [x index={0},y index={1}]{pr_dif_detectors_brisk_akaze_orb_128_firewire.txt};
		\addplot+[mark=+,color=black,line width=1pt]
		table [x index={2},y index={3}]{pr_dif_detectors_brisk_akaze_orb_128_firewire.txt};
		\addplot+[mark=+,color=orange,line width=1pt]
		table [x index={4},y index={5}]{pr_dif_detectors_brisk_akaze_orb_128_firewire.txt};
		\addplot+[mark=+,color=green,line width=1pt]
		table [x index={6},y index={7}]{pr_dif_detectors_brisk_akaze_orb_128_firewire.txt};
		\addplot+[mark=+,color=red,line width=1pt]
		table [x index={8},y index={9}]{pr_dif_detectors_brisk_akaze_orb_128_firewire.txt};
		\addplot+[mark=o,color=green,line width=1pt]
		table [x index={10},y index={11}]{pr_dif_detectors_brisk_akaze_orb_128_firewire.txt};
		\addplot+[mark=o,color=red,line width=1pt]
		table [x index={12},y index={13}]{pr_dif_detectors_brisk_akaze_orb_128_firewire.txt};			
		\end{axis}
		\end{tikzpicture}
		\includegraphics[width=0.41\textwidth]{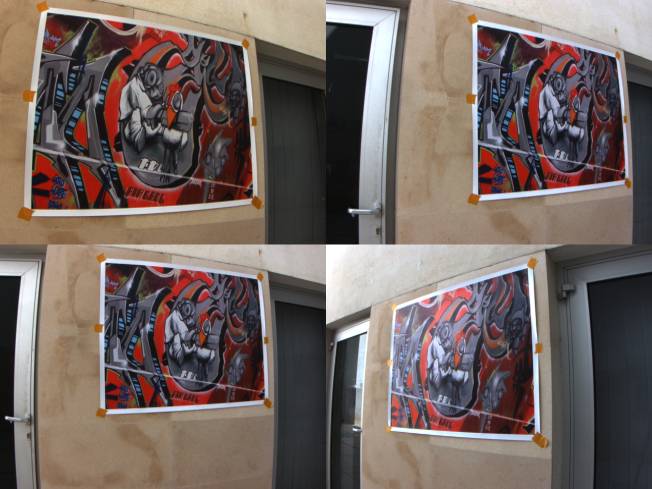}
	}\\
	\subcaptionbox{Dataset Dragonfly. \label{fig:medDistortionSRD}}
	{
		\begin{tikzpicture}[baseline=(current axis.south)]
		\begin{axis}[
		grid=major,
		ymax=1.0,ymin=0,
		xmax=0.7,xmin=0,
		xtick={0.2,0.4,0.6,0.8},
		xlabel={1-precision},
		ylabel={Recognition rate (recall)},
		yticklabel pos=left,
		width = 0.5\textwidth,
		height = 0.4\textwidth,
		ylabel near ticks,
		xlabel near ticks]
		\addplot+[mark=+,color=blue,line width=1pt]
		table [x index={0},y index={1}]{pr_dif_detectors_brisk_akaze_orb_128_dragonfly.txt};
		\addplot+[mark=+,color=black,line width=1pt]
		table [x index={2},y index={3}]{pr_dif_detectors_brisk_akaze_orb_128_dragonfly.txt};
		\addplot+[mark=+,color=orange,line width=1pt]
		table [x index={4},y index={5}]{pr_dif_detectors_brisk_akaze_orb_128_dragonfly.txt};
		\addplot+[mark=+,color=green,line width=1pt]
		table [x index={6},y index={7}]{pr_dif_detectors_brisk_akaze_orb_128_dragonfly.txt};	
		\addplot+[mark=+,color=red,line width=1pt]
		table [x index={8},y index={9}]{pr_dif_detectors_brisk_akaze_orb_128_dragonfly.txt};
		\addplot+[mark=o,color=green,line width=1pt]
		table [x index={10},y index={11}]{pr_dif_detectors_brisk_akaze_orb_128_dragonfly.txt};
		\addplot+[mark=o,color=red,line width=1pt]
		table [x index={12},y index={13}]{pr_dif_detectors_brisk_akaze_orb_128_dragonfly.txt};				
		\end{axis}
	\end{tikzpicture}
	\includegraphics[width=0.41\textwidth]{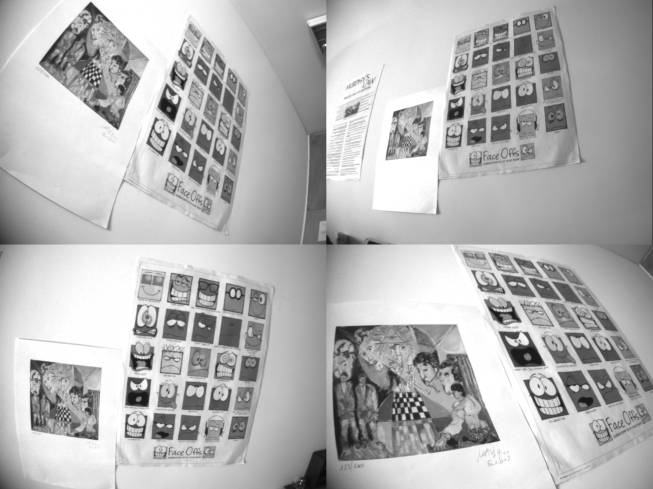}
	} \\
	\subcaptionbox{Dataset Fisheye. \label{fig:highDistortionSRD}}
	{
		\begin{tikzpicture}[baseline=(current axis.south)]
		\begin{axis}[
		grid=major,
		ymax=1.0,ymin=0,
		xmax=0.7,xmin=0,
		xtick={0.2,0.4,0.6,0.8},
		xlabel={1-precision},
		ylabel={Recognition rate (recall)},
		yticklabel pos=left,
		width = 0.5\textwidth,
		height = 0.4\textwidth,
		ylabel near ticks,
		xlabel near ticks]
		\addplot+[mark=+,color=blue,line width=1pt]
		table [x index={0},y index={1}]{pr_dif_detectors_brisk_akaze_orb_128_firewireFish.txt};
		\addplot+[mark=+,color=black,line width=1pt]
		table [x index={2},y index={3}]{pr_dif_detectors_brisk_akaze_orb_128_firewireFish.txt};
		\addplot+[mark=+,color=orange,line width=1pt]
		table [x index={4},y index={5}]{pr_dif_detectors_brisk_akaze_orb_128_firewireFish.txt};
		\addplot+[mark=+,color=green,line width=1pt]
		table [x index={6},y index={7}]{pr_dif_detectors_brisk_akaze_orb_128_firewireFish.txt};
		\addplot+[mark=+,color=red,line width=1pt]
		table [x index={8},y index={9}]{pr_dif_detectors_brisk_akaze_orb_128_firewireFish.txt};
		\addplot+[mark=o,color=green,line width=1pt]
		table [x index={10},y index={11}]{pr_dif_detectors_brisk_akaze_orb_128_firewireFish.txt};
		\addplot+[mark=o,color=red,line width=1pt]
		table [x index={12},y index={13}]{pr_dif_detectors_brisk_akaze_orb_128_firewireFish.txt};		
		\end{axis}
		\end{tikzpicture}
		\includegraphics[width=0.41\textwidth]{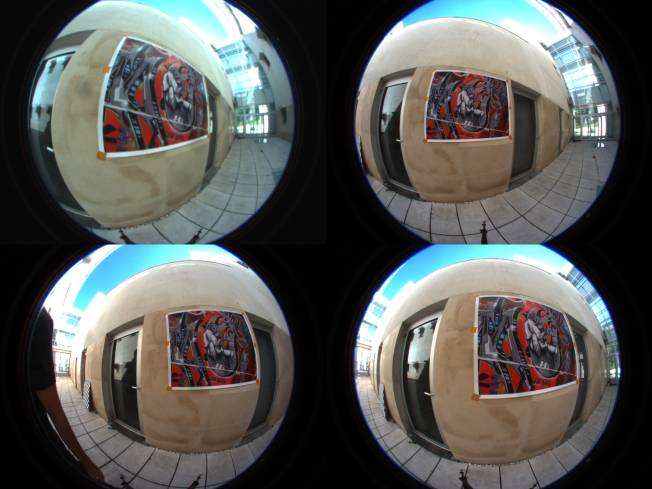}
	} 
	\begin{center}
		\ref{legendPRCurves}
	\end{center}	
	\caption{sRD-SIFT data set. For each image we extract the 300 strongest keypoints. For BRISK, ORB and AKAZE we use the standard settings. The BOLD descriptor is extracted on the same patches as the mdBRIEF descriptor.}
	\label{fig:srdsiftPR}
\end{figure*}

\paragraph{sRD-SIFT data set}
The second dataset was published with the work on sRD-SIFT \cite{lourencco2012srd}.
It consists of three image sets of 13 images respectively.
Each image set was recorded with cameras having different amount of radial distortion.
In addition, the scene contains significant scale and rotation change.
Four randomly chosen images of every dataset are depicted in \Seefig{fig:srdsiftPR}.

The authors use a different camera model, but attached the calibration images to the data set.
Thus, we calibrate each camera using the camera model and calibration procedure introduced in \cite{urban2015improved} using the provided checkerboard images.
The ground truth homographies are estimated according to the method used for the fisheye sequence dataset.

Again, we perform Brute Force descriptor matching and select ground truth matches with a maximal distance of three pixels.
In addition, descriptors are extracted with and without patch orientation as two of the datasets contained only little rotation about the camera  view direction.
The PR-curves are depicted in \Seefig{fig:srdsiftPR}.
\Seefig{fig:lowDistortionSRD} shows the results for the image set with the smallest amount of distortion.
\Seefig{fig:medDistortionSRD} shows the results for the image set with a medium amount of distortion.
\Seefig{fig:highDistortionSRD} depicts the results for the fisheye data set.
Here, we extracted a mask of the planar Graffiti region along with the ground truth homographies.
The Graffiti poster occupies only a small image region and most features were detected at the floor and the doors in the right and left part of the images.
Those, however, are not part of the Graffiti plane, and can hence not be transformed with the ground truth homographies and are marked as false matches.
\paragraph{Extraction Speed}
As indicated in the introduction of this chapter, the total amount of time to detect, describe and match keypoints is important to evaluate if the application of a specific detector-desciptor combination is practicable.
Our two descriptor variants dBRIEF and mdBRIEF are implemented in C++ and can be used as an add-on to OpenCV.
Thus, the timings are directly comparable to the OpenCV implementations.
For mdBRIEF, the mask learning is performed in parallel on 4 threads.
\SeeTable{tab:extratctionSpeed} shows timings for most state-of-the-art detector-descriptor combinations that are available through OpenCV.
Obviously, the mask learning as well as the distortion of the test set has its price.
But still, the combination of dBRIEF and mdBRIEF with the fast ORB detector  leaves us with real time capable timings, apart from being more accurate than other detector-descriptor combinations as shown in the previous section.
From \SeeTable{tab:extratctionSpeed} the additional effort of matching masked descriptors is visible.
\begin{table*}
	\centering
\begin{filecontents*}{results_timing.csv}
Detector,Descriptor,Detection,Description,Matching,Total
ORB, BRIEF, 13.1407, 3.7402, 4.14022, 16.8851
ORB, ORB, 14.5208, 8.22046, 4.04022, 22.7453
modORB, dBRIEF, 6.40036, 17.941, 4.20024, 24.3456
BRISK, BRISK, 15.2363, 10.3191, 7.47236, 25.5629
modORB, mdBRIEF, 6.46036, 45.2626, 8.64048, 51.7316
ORB, BOLD, 13.3808, 43.2825, 12.5007, 56.6757
SURF, SURF, 25.6372, 38.8898, 54.6343, 64.5815
ORB, LATCH, 13.0607, 58.6233, 4.00022, 71.6881
AKAZE, AKAZE, 75.2293, 55.0298, 6.83104, 130.266
SIFT, SIFT, 106.208, 133.615, 114.638, 239.938
\end{filecontents*}
\pgfplotstabletypeset[
columns/Detector/.style={string type},
columns/Descriptor/.style={string type},
columns/{Detection}/.style={column name={Detection [$\mu$s]},precision=0},
columns/{Description}/.style={column name={Description [$\mu$s]},precision=0},
columns/{Matching}/.style={column name={Matching [ns]},precision=0},
columns/{Total}/.style={column name={Total [$\mu$s]},precision=0},
col sep=comma,
every even row/.style={before row={\rowcolor[gray]{0.9}}},
every head row/.style={before row=\toprule,after row=\midrule},
every last row/.style={after row=\bottomrule}]
{results_timing.csv}
\caption{Extraction speed. Comparison of state-of-the-art methods. Implementations are based on OpenCV 3.1. We use the 256 bit variants of dBRIEF and mdBRIEF.}
\label{tab:extratctionSpeed}
\end{table*}
\section{Summary}
\label{ch:features:sec:conclusion}
In this chapter, a modified version of the BRIEF descriptor was introduced.
At the beginning, basic concepts of keypoint detection, patch description and matching were recalled.
Then, the current state-of-the-art was analyzed and recapitulated.
From this analysis it was possible to deduce possible detector-descriptor combinations for the task at hand.
As the given methods were basically developed for perspective cameras, we combined the advantages of BRIEF, ORB and BOLD to come up with a distorted version of a binary test based descriptor.
Various simulations emphasized the advantages of using the interior orientation of the camera to distort the binary test set.
Finally the developed dBRIEF and mdBRIEF descriptors were tested on real data and the performance was evaluated in terms of precision and recall.
Both descriptors show good results and outperform other state-of-the-art descriptors, whilst being real-time capable.

\section*{Acknowledgment}
This project was partially funded by the DFG research group FG 1546 "Computer-Aided Collaborative Subway Track Planning in Multi-Scale 3D City and Building Models".
\begin{spacing}{0.8}
	{\small \bibliography{mdBrief_bib}} 
\end{spacing}

\end{document}